\documentclass[12pt,a4paper]{article}

\usepackage{geometry}

\usepackage{amsfonts}
\usepackage{amsthm}
\usepackage{graphicx}
\usepackage{url}

\usepackage{multicol}
\usepackage{lipsum}

\usepackage{amsmath}
\usepackage{amssymb}
\usepackage{bm}
\usepackage{tcolorbox}

\newtheorem{proposition}{Proposition}

\usepackage[normalem]{ulem}
\usepackage{xcolor}

\usepackage{adjustbox}
\usepackage{tikz}
\usetikzlibrary{arrows.meta, backgrounds, decorations, fit, positioning, calc}
\tikzset{fit margins/.style={/tikz/afit/.cd,#1,
    /tikz/.cd,
    inner xsep=\pgfkeysvalueof{/tikz/afit/left}+\pgfkeysvalueof{/tikz/afit/right},
    inner ysep=\pgfkeysvalueof{/tikz/afit/top}+\pgfkeysvalueof{/tikz/afit/bottom},
    xshift=-\pgfkeysvalueof{/tikz/afit/left}+\pgfkeysvalueof{/tikz/afit/right},
    yshift=-\pgfkeysvalueof{/tikz/afit/bottom}+\pgfkeysvalueof{/tikz/afit/top}},
    afit/.cd,left/.initial=2pt,right/.initial=2pt,bottom/.initial=2pt,top/.initial=2pt}

\usepackage[pagebackref=true,breaklinks=true,letterpaper=true,colorlinks,bookmarks=false]{hyperref}

\newcommand{\mE}{{\mathbb{E}}}
\newcommand{\sP}{{\cal P}}
\newcommand{\sD}{{\cal D}}
\def\w{{\bf w}}
\newcommand{\bc}{\begin{center}}
\newcommand{\ec}{\end{center}}
\newcommand{\bit}{\begin{itemize}}
\newcommand{\eit}{\end{itemize}}
\newcommand{\ben}{\begin{enumerate}}
\newcommand{\een}{\end{enumerate}}
\newcommand{\bH}{{H}}
\def\x{{x}}
\def\X{{X}}
\def\tx{\tilde{x}}
\def\z{{z}}
\DeclareMathOperator\supp{supp}
\newcommand{\diff}{\mathop{}\!d}

\newtheorem{theorem}{Theorem}

\newcommand{\myskip}[1]{ }

\newcommand\blfootnote[1]{%
  \begingroup
  \renewcommand\thefootnote{}\footnote{#1}%
  \addtocounter{footnote}{-1}%
  \endgroup
}

\title{A Causal Framework to Unify \\ Common Domain Generalization Approaches }

\author{
Nevin L. Zhang\textsuperscript{\rm 1,$*$}
~
Kaican Li\textsuperscript{\rm 1,$*$}
~
Han Gao\textsuperscript{\rm 2,$*$} 
~
Weiyan Xie
~
Zhi Lin
\\[0.1cm]
~
Zhenguo Li
~
Luning Wang
~
Yongxiang Huang
\\ [0.25cm]
\textsuperscript{\rm 1}The Hong Kong University of Science and Technology
\\
\textsuperscript{\rm 2} Huawei Hong Kong AI Framework \& Data Technologies Lab
\\[0.1cm]
{$^1$ \tt\small \{klibf, lzhang\}@cse.ust.hk}
~
{$^2$ \tt\small gaohan19@huawei.com}
}

\date{}

\begin{document}

\maketitle

\begin{abstract}
Domain generalization (DG) is about learning models that generalize well to new domains that are related to, but different from, the training domain(s). It is a fundamental problem in machine learning and has attracted much attention in recent years.
A large number of approaches have been proposed. Different approaches are motivated from different perspectives, making it difficult to gain an overall understanding of the area.  In this paper, we propose a causal framework for domain generalization and present an  understanding of common DG approaches in the framework. Our work sheds new lights on the following questions: (1) What are the key ideas behind each DG method? (2)
Why is it expected to improve generalization to new domains theoretically? (3)
How are different DG methods  related to each other and what are relative advantages and limitations?  By providing a unified perspective on DG, we hope to help researchers better understand the underlying principles and develop more effective approaches for this critical problem in machine learning.

\blfootnote{$*$ Equal contribution.}
\vspace{-4mm}
\end{abstract}

\section{Introduction}

The standard supervised  learning paradigm involves three datasets: a training set, a validation set, and a test set. The training set is used to optimize the parameters of a model, the validation set is used to choose the hyperparameters, and the test set is used to evaluate the performance of the model. The model will be applied in the future by  different users to different collections of data, which are called {\em target sets}. The training, validation and test sets are independent and identically distributed (i.i.d.)\ by design. However, a target set usually follows a different distribution. In this sense
there is {\em domain shift} between the training and target domains.
Model accuracy on
the test set is known as the  {\em in-distribution (ID) accuracy}, and model accuracy on a target set is known as the {\em out-of-distribution (OOD) accuracy}.

The performance of machine learning models is typically assessed using the ID accuracy. However,
it has been observed that, due to domain shift, the OOD accuracy is almost always significantly lower than the ID
accuracy. For example, Taori {\em et al}.~\cite{taori2020measuring} took 204 models that were trained on the
ImageNet dataset~\cite{deng2009imagenet} and tested them on ImageNet-V2~\cite{recht2019imagenet}, another dataset created in the same
way as ImageNet except one decade later. They found that the OOD accuracy is more than 10\% lower than the ID
accuracy for all the models.
Beery {\em et al.}~\cite{beery2018recognition} demonstrated that an  object recognition model trained on images of
cows appearing on green pastures fails to recognize a cow on a beach.
Alcorn {\em et al.}~\cite{alcorn2019strike} showed that  an image classifier can incorrectly classify school buses not in their canonical pose as in the training set.
Zech {\em et al}.~\cite{zech2018variable} and
Degrave {\em et al.}~\cite{degrave2021ai} trained models
for detecting
pneumonia and COVID-19  in chest radiographs using data from some hospitals. They found that the models perform much worse
on external test sets from other hospitals.

\begin{center}
	\begin{figure}[t]
		\begin{center}
			\small
			\begin{tabular}{cp{0.5cm}c}
				\includegraphics[width=8cm]{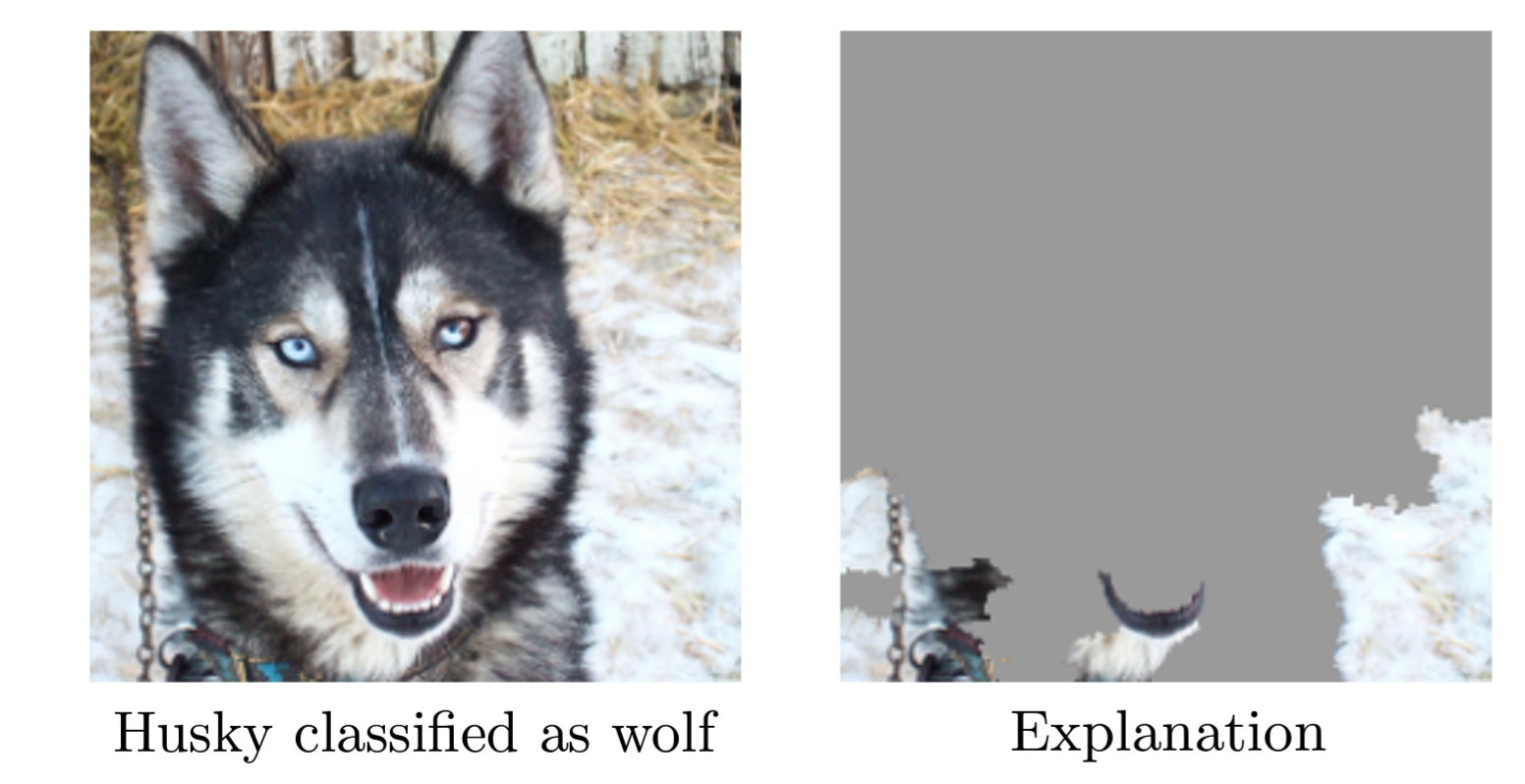}  &  &	\includegraphics[width=3.6cm]{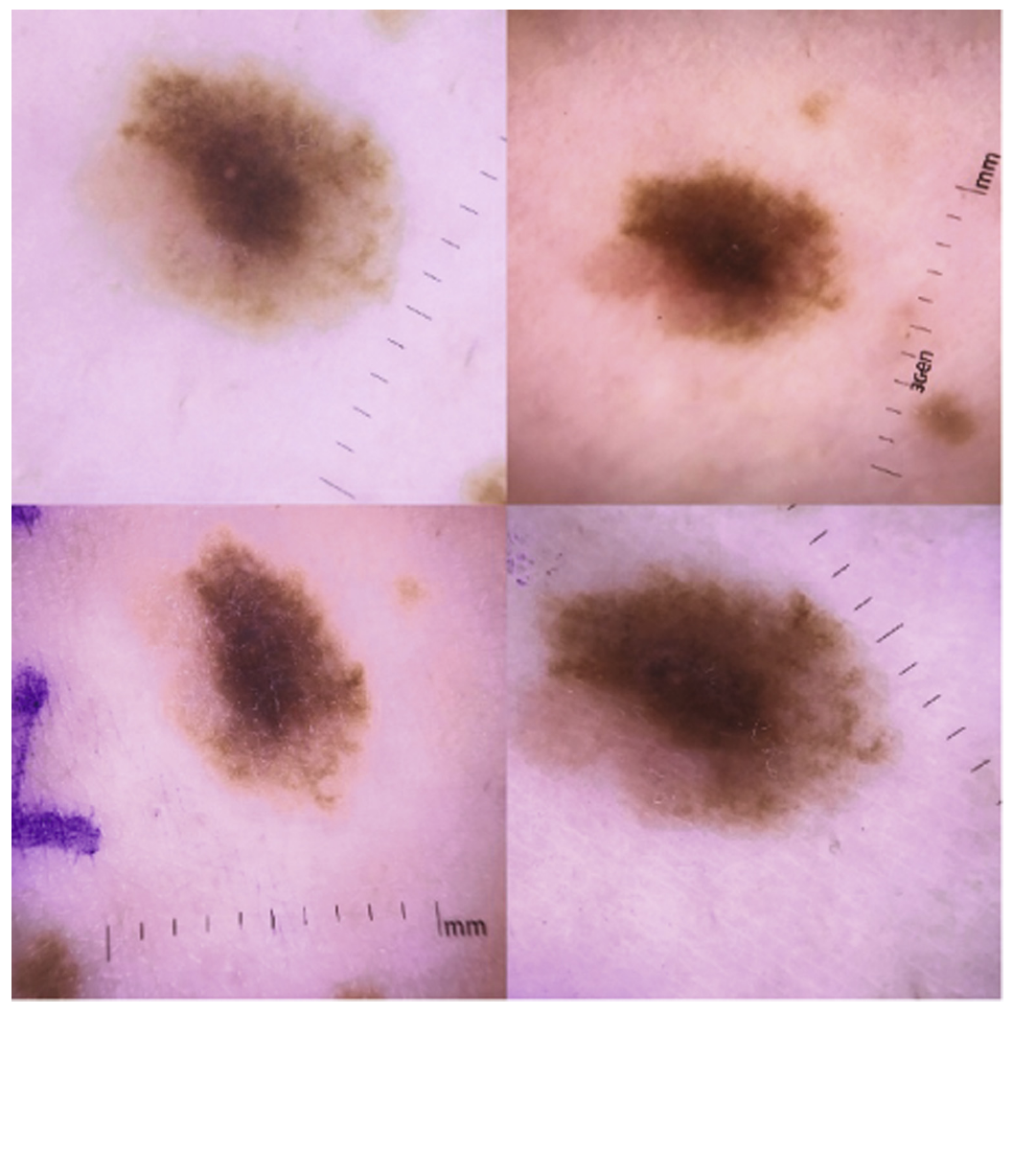} \\
				\small	(a)  & & \small (b)
				
			\end{tabular}
		\end{center}
		\caption{\small Deep neural network models tend to exploit  spurious correlations: (a) Snowy backgrounds are associated with wolves~\cite{ribeiro2016should}; (b) rulers are associated with malignant skin lesions~\cite{narla2018automated}.  }
		\label{fig.spurious}
	\end{figure}
\end{center}

Domain shift leads to poor  generalization to new domains because deep neural network models tend to exploit {\em spurious correlations}, i.e., correlations between  input and output that hold in the source domain but not in the target domain. Figure~\ref{fig.spurious} provides two examples of spurious correlations. Subfigure (a) shows a model trained on a dataset where all pictures of wolves had snow in the background, while pictures of huskies did not. The model learned to associate snowy backgrounds with wolves, which is true in the training set but not in general,  leading it to misclassify a husky in snow as a
wolf~\cite{ribeiro2016should}.
Subfigure (b) demonstrates how a skin lesion classification model \cite{narla2018automated} incorrectly associates rulers with malignant skin lesions just because images with rulers were more likely to be malignant in the training set.

{\em Domain generalization (DG)}, also known as OOD generalization, refers to the task of reducing the impact of
spurious correlations and making models generalize well to unseen domains.
 This problem has attracted significant attention in the past decade, leading to a vast literature on DG that can be formidable for readers. To aid in navigating this literature, several comprehensive survey
 papers~\cite{shen2021towards,zhou2021domain,sheth2022domain,wang2022generalizing} have recently appeared. These papers typically categorize DG algorithms into several classes and briefly discuss several algorithms in each category.
For example, Wang  {\em et al.}~\cite{wang2022generalizing} divide DG methods into three categories:
data manipulation, representation learning, and learning strategy. They cover a large number of specific methods, including data augmentation, feature alignment, domain adversarial learning, invariant risk minimization, causality-inspired methods, meta-learning, and gradient manipulation.   However, since different methods are motivated from different perspectives, it can be challenging to understand  how they are related to each other and what their relative strengths and weaknesses are.  Furthermore, many methods lack a solid theoretical foundation, leaving it difficult to pinpoint the exact reason why they can improve generalization to new domains.

This paper proposes a causal framework for domain generalization (DG) that unifies common DG methods. We first establish a set of sufficient conditions for optimal generalization to new domains.
We then explain common DG methods are various ways to achieve those conditions directly or indirectly, or to achieve some alternative conditions. 
Our work sheds new lights on the following questions: (1) What are the key ideas behind each DG method? (2)
Why is it expected to improve generalization to new domains theoretically? (3)
How are different DG methods  related to each other and what are relative advantages and limitations?

The rest of this paper is organized as follows.
 Section \ref{sec.model} presents a causal model for DG and  introduces the key concepts about DG. Section \ref{sec.conditions} establishes theoretical conditions for optimal domain generalization. Sections \ref{sec.pair}-\ref{sec.single-source} explain how common DG methods are related to those conditions, with a summary given in Section \ref{sec.summary}.
 Finally,  we end the paper at Section \ref{sec.conclusion} with a few concluding remarks.

\section{A Causal Framework for Domain Generalization}
\label{sec.model}

In domain generalization, a {\em domain} $d$ is defined by a distribution $P^d(\X, Y)$
over the space of input-label pairs $(\X, Y)$. This paper focuses on the case where $\X$ is an image and $Y$ is a class label.  The general DG formulation involves a
family ${\cal P}$ of domains.  It assumes training data from one or multiple domains from ${\cal P}$, which are called  {\em source domains}.  The task is to learn a prediction model that performs well not only in the source domains but also in all the other domains in ${\cal P}$, which are the {\em target domains}.

\subsection{Model for Data Generation}

\label{sec.dataModel}

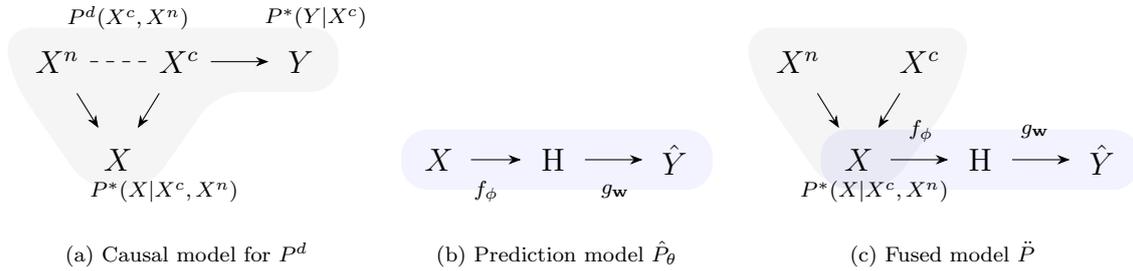
\begin{figure}[t]
\centering
	\begin{tabular}{ccc}
		\begin{tikzpicture}
			\begin{scope}[every node/.style={minimum size=2em}]
				\node (xn0) at (0, 0) {$\X^n$};
				\node (xc0) at (1.6, 0) {$\X^c$};
				\node (y)  at (3.2, 0) {$Y$};
				\node (x) at (0.8, -1.3) {$\X$};\
                \draw (0.1, 1.0) node [anchor=north west][inner sep=0.75pt]  [font=\scriptsize] [align=left] {$P^{d}(X^{c}, X^{n})$};
                \draw (0.4, -1.3) node [anchor=north west][inner sep=0.75pt]  [font=\scriptsize] [align=left] {$P^{*}(X|X^{c}, X^{n})$};
                \draw (2.7, 1.0) node [anchor=north west][inner sep=0.75pt]  [font=\scriptsize] [align=left] {$P^{*}(Y|X^{c})$};
			\end{scope}
			\begin{scope}[>={Stealth[black]},
				every edge/.style={draw=black}]
				\path [dashed] (xn0) edge node {} (xc0);
				\path [->] (xc0) edge node {} (y);
				\path [->] (xn0) edge node {} (x);
				\path [->] (xc0) edge node {} (x);
			\end{scope}
			\begin{scope}[on background layer]
				\draw[rounded corners=10, fill=black, fill opacity=0.04, draw=none]($(xn0) + (-0.6, -0.25)$) -- ($(xn0) + (-0.7, 0.45)$) -- ($(y) + (0.5, 0.45)$)  -- ($(y) + (0.5, -0.4)$) -- ($(xc0) + (0.5, -0.4)$) -- ($(x) + (0.4, -0.4)$) -- ($(x) + (-0.4, -0.4)$) -- cycle;
			\end{scope}
		\end{tikzpicture}
		& 	\begin{tikzpicture}
			\begin{scope}[every node/.style={minimum size=2em}]
				\node (yhat)  at (3.4, 0.3) {$\hat{Y}$};
				\node (x) at (0.3, 0.3) {$\X$};
                \node (H) at (1.8, 0.3) {$\textup{H}$};
                \draw (0.5, 0.3) node [anchor=north west][inner sep=0.75pt]  [font=\scriptsize] [align=left] {$f_{\phi}$};
                \draw (2.2, 0.3) node [anchor=north west][inner sep=0.75pt]  [font=\scriptsize] [align=left] {$g_{\textbf{w}}$};
			\end{scope}
			\begin{scope}[>={Stealth[black]},
				every edge/.style={draw=black}]
				\path [->] (x) edge node {} (H);
                \path [->] (H) edge node {} (yhat);
			\end{scope}
			\begin{scope}[on background layer]
				\draw[rounded corners=10, fill=blue, fill opacity=0.05, draw=none]($(x) + (-0.5, -0.4)$) -- ($(x) + (-0.5, 0.4)$) -- ($(yhat) + (0.5, 0.4)$)  -- ($(yhat) + (0.5, -0.4)$) -- cycle;
			\end{scope}
		\end{tikzpicture}
		& 	\begin{tikzpicture}
			\begin{scope}[every node/.style={minimum size=2em}]
				\node (xn0) at (0, 0) {$\X^n$};
				\node (xc0) at (1.6, 0) {$\X^c$};
				\node (yhat)  at (4.0, -1.3) {$\hat{Y}$};
				\node (x) at (0.8, -1.3) {$\X$};
                \node (H) at (2.4, -1.3) {$\textup{H}$};
                \draw (0.0, -1.3) node [anchor=north west][inner sep=0.75pt]  [font=\scriptsize] [align=left] {$P^{*}(X|X^{c}, X^{n})$};
                \draw (1.2, -0.5) node [anchor=north west][inner sep=0.75pt]  [font=\scriptsize] [align=left] {$f_{\phi}$};
                \draw (2.7, -0.5) node [anchor=north west][inner sep=0.75pt]  [font=\scriptsize] [align=left] {$g_{\textbf{w}}$};
			\end{scope}
			\begin{scope}[>={Stealth[black]},
				every edge/.style={draw=black}]
				\path [->] (xn0) edge node {} (x);
				\path [->] (xc0) edge node {} (x);
				\path [->] (x) edge node {} (H);
                \path [->] (H) edge node {} (yhat);
			\end{scope}
			\begin{scope}[on background layer]
				\draw[rounded corners=10, fill=black, fill opacity=0.04, draw=none]($(xn0) + (-0.6, -0.25)$) -- ($(xn0) + (-0.7, 0.45)$) -- ($(xc0) + (0.7, 0.45)$) -- ($(xc0) + (0.5, -0.4)$) -- ($(x) + (0.4, -0.4)$) -- ($(x) + (-0.4, -0.4)$) -- cycle;
				\draw[rounded corners=10, fill=blue, fill opacity=0.05, draw=none]($(x) + (-0.5, -0.4)$) -- ($(x) + (-0.5, 0.4)$) -- ($(yhat) + (0.5, 0.4)$)  -- ($(yhat) + (0.5, -0.4)$) -- cycle;
			\end{scope}
		\end{tikzpicture} \\
		\scriptsize (a) Causal model for $P^{d}$  & \scriptsize  (b) Prediction model $\hat{P}_{\theta}$  & \scriptsize (c) Fused model $\ddot{P}$
	\end{tabular}
	\caption{\small A causal DG framework: (a).\ A data generation model for a domain $P^d(\X^c, \X^n, \X, Y)$. It is called the CLD model, where $P^*(\X|X^c, \X^n)$ and $P^*(Y|\X^c)$ are invariant across domains, as indicated by the solid edges. (b).\ A prediction model $\hat{P}_{\theta}(\hat{Y}|\X)$ is trained on data from one or multiple training domains. It consists of a feature extractor $f_{\phi}$ and a linear classification head $g_{\w}$. $H = f_{\phi}(\X)$ is the vector of features extracted from input $X$.  (c).\
		$P^*(\X|X^c, \X^n)$ and $\hat{P}_{\theta}(\hat{Y}|\X)$   are combined to form a fused model $\ddot{P}_{\theta}$, which enables us to discuss the dependence of prediction output $\hat{Y}$ on the latent factors behind inputs.
		Note that in (1) $\X^c$ is the cause of $y$ in $P^d$ while $\X$ is the cause of $\hat{Y}$ in  $\hat{P}_{\theta}$ and $\ddot{P}_{\theta}$, and (2) $\X \perp Y|\X^c$ in $P^d$, while $\X^c \perp \hat{Y}|\X$ in $\ddot{P}$.
	}
	\label{fig.causalModel}
\end{figure}

In this paper we assume that each domain distribution $P^d(\X, Y)$ in a family $\sP$ is a marginal of another distribution $P^d(\X^c, \X^n, \X, Y)$ that involves two aditional latent variables $\X^c$ and $\X^n$.  The qualitative relationships among the four variables
are depicted as a graphical model in Fig.~\ref{fig.causalModel} (a).
The model assumes that: (1) an example input $\X$ is generated from two latent variables
 $\X^c$ and $\X^n$, (2) $\X^c$ and $\X^n$ are statistically correlated, and (3) the label $Y$ is generated from only $\X^c$.
 The {\em causal mechanisms} that generate $\X$ and $Y$ are assumed to be invariant across domains \cite{hoover1990logic,zhang2013domain,peters2016causal}. This is known as  the
 {\em causal invariance assumption}.
 The correlation between $\X^n$ and $\X^c$ may be due to: (1) a direct causal link from $\X^c$ to $\X^n$, (2) a direct causal link from $\X^n$ to $\X^c$, (3) exogenous confounding factors that causes both $\x^c$ and $\X^n$, and (4) a combination of 1-3 and other reasons.  It  changes from one domain to another.

The  graphical model introduced above first appeared in~\cite{tenenbaum1996separating}, where it is called the {\em style and content decomposition (SCD) model}, and
$\X^c$ and $\X^n$ are called the  {\em content} and {\em style} variables
respectively.
Similar models  appeared recently in a number of papers under different terminologies, which are summarized in Table \ref{table.terms}.  {\em  The  variable $\X^c$ denotes the essential
	 information in an image $\X$  that a human relies on to
	assign a label $Y$ to the image.}  It is hence said to represent
{causal factors}~\cite{lv2022causality}, {intended  factors}~\cite{geirhos2020shortcut}, {semantic factors}~\cite{liu2021learning},
{content factors}~\cite{mitrovic2021representation}, and
{core factors}~\cite{heinze2021conditional}.  In contrast,
{\em the variable $\X^n$
	denotes the other aspects of $\X$ that are not essential to label assignment.} It
is hence said to represent
{non-causal factors}, {non-intended  factors}, {variation factors},  {style factors}, and
{non-core factors}.
As the relationship between $\X^c$ and $Y$ does not change across domains,
 $\X^c$ is sometimes said to represent
{stable features}~\cite{zhang2021deep},
domain-independent factors~\cite{ouyang2022causality},
and invariant features~\cite{arjovsky2019invariant,ahuja2021invariance}.
In contrast, $\X^n$ is said to represent
non-stable features,
domain-dependent factors,
and spurious features.


	\begin{table}[t]
		\caption{Different authors refer to the two latent variables $\X^c$ and $\X^n$ using different terms, which provide multiple perspectives on their semantics.    }
	 \label{table.terms}
	
	\begin{center}
		\begin{tabular}{|c||l|l|l|}
			\hline
			$\X^c$ & \small content factors~\cite{mitrovic2021representation} & \small semantic factors~\cite{liu2021learning} & \small
			core factors~\cite{heinze2021conditional} \\
			$\X^n$ & \small style factors & \small variation  factors & \small non-core  factors \\ \hline
			$\X^c$ &  \small {causal factors~\cite{lv2022causality}} & \small intended  factors~\cite{geirhos2020shortcut} & \\
			$\X^n$ &  \small non-causal factors & \small non-intended  factors & \\ \hline
			$\X^c$ & \small stable features~\cite{zhang2021deep} & \small  domain-independent factors~\cite{ouyang2022causality} &  \small invariant features~\cite{arjovsky2019invariant} \\
			$\X^n$ & \small non-stable features& \small domain-dependent factors& \small spurious features\\
			\hline
		\end{tabular}
		
	\end{center}
		\end{table}

The term ``style" in the SCD model should be understood in a broad sense. In additional to image style,  it also includes factors such as background, context, object pose and so on.  To avoid ambiguity, we follow ~\cite{heinze2021conditional} and 
refer to
$\X^c$ and $\X^n$ as the {\em core and non-core  factors} respectively,
and rename the SCD model as the {\em causal latent decomposition (CLD) model}.
Besides latent decomposition,  the model  also encapsulates the causal invariance assumption regarding data generation. We will later introduce a similar concept for prediction.  For clarity, we will refer to the assumption here
is the {\em causal-invariant generation (CIG) assumption}.

To ground the CLD model, we need to specify three distributions: $P^d(\X^c, \X^n)$, $P^d(\X|\X^c, \X^n)$ and $P^d(Y|\X^c)$.  Due to the CIG assumption, the last two distributions do not change across domains. We denote them as
$P^*(\X|\X^c, \X^n)$ and $P^*(Y|\X^c)$.  In contrast, $P^d(\X^c, \X^n)$ may be different for different domains.   Together, the three distributions define a joint distribution over the four variables
\begin{eqnarray*}
	P^d(\X^c, \X^n, \X, Y) = P^d(\X^c, \X^n)P^*(\X|\X^c, \X^n)P^*(Y|\X^c).
\end{eqnarray*}
This joint distribution defines a domain in the CLD framework.  We refer to the collection of all such domains for some fixed $P^*(\X|\X^c, \X^n)$ and $P^*(Y|\X^c)$ as a
 {\em CLD family}:
 \[\sP_{P^*(\X|\X^c, \X^n),P^*(Y|\X^c)} = 
 \{P^d(\X^c, \X^n, \X, Y) | P^d(\X^c, \X^n)\}.\]
 In each domain $d$, examples are  generated as follows:
\begin{eqnarray*}
	(\x^c, \x^n) &\sim&  P^d(\X^c, \X^n) \\
	\x & \sim & P^*(\X|\x^c, \x^n)\\
	y & \sim &P^*(Y|\x^c), 
\end{eqnarray*}
\noindent where $\x$ and $y$ are observed while $\x^n$ and $\x^c$ are hidden.
Note that we will use upper-case letters such as $\X$ and $Y$ to denote variables, and use lower case letters such as $\x$ and $y$ to denote their values. $P^d(\X^c, \X^n)$ denotes the joint distribution of $\X^c$ and $\X^n$,  $P^*(\X|\x^c, \x^n)$ is the conditional distribution of $X$ given $\X^c = \x^c$ and $\X^n=\x^n$, and $P^*(Y|\x^c)$ is the conditional distribution of $Y$ given $\X^c = \x^c$.

\subsection{Domain Shifts and Spurious Correlations}

Consider a source domain $P^s$ and a target domain  $P^t$   from a CLD family.
Usually $P^{s}(\X^c, \X^n)$ differs from $P^{t}(\X^c, \X^n)$, which may lead to:
\begin{eqnarray*} P^{s}(\X) \neq P^{t}(\X), \hspace{0.3cm} P^{s}(Y|\X) \neq P^{t}(Y|\X), \hspace{0.3cm} P^{s}(Y|\X^n) \neq P^{t}(Y|\X^n).
\end{eqnarray*}
\noindent The first inequality states that different domains may have different distributions of training examples. This is known as
{\em covariate shift}~\cite{zhang2013domain}.  The second inequality
means that the conditional distribution of $Y$ given $\X$ may change from one domain to another. This is known as {\em concept shift}~\cite{zhang2013domain}.
The third inequality states that the correlations between $Y$ and $\X^n$ may vary across domains. In this sense those correlations are {\em spurious} \cite{liu2021heterogeneous}.

Domain generalization in the presence of both covariate shift and concept shift is difficult.  Some assumptions must be made.  In this paper we adopt the CIG  assumption, i.e., $P^{*}(\X|\X^c, \X^n)$ and  $P^{*}(Y|\X^c)$ remain constant across domains.

\subsection{Prediction Model and Fused Model}
\label{sec.fusedModel}

Besides a data generation model, our framework involves two other models: a prediction model  and a fused model.
A {\em prediction model}  $\hat{P}_{\theta}(\hat{Y}|\X)$
(Fig.~\ref{fig.causalModel} (b)) is a neural network model with parameters $\theta$.
It is trained on a source domain or multiple source domains.
The {\em fused model}  $\ddot{P}$ (Fig.~\ref{fig.causalModel}  (c)) is a technical model created by combining the prediction model $\hat{P}_{\theta}(\hat{Y}|\X)$ and $P^*(\X|\X^c, \X^n)$:
\begin{eqnarray}
	\label{eq.fused}
	\ddot{P}_{\theta}(\X, \hat{Y}|\X^c, \X^n)  = \hat{P}_{\theta}(\hat{Y}|\X)P^*(\X|\X^c, \X^n).
\end{eqnarray}
The two dots  on top of $\ddot{P}_{\theta}$ indicate the combination of two models. The fused model  enables us to characterize the dependence of prediction output $\hat{Y}$ on the latent factors $\X^c$ and $\X^n$ behind the input  $\X$.
 Note that  $\X^c$ causes $Y$ in $P^d$, while $\X$ causes $\hat{Y}$ in  $\hat{P}_{\theta}$ and $\ddot{P}_{\theta}$. Furthermore,  $\X \perp Y|\X^c$ in $P^d$, while $\X^c \perp \hat{Y}|\X$ in $\ddot{P}_{\theta}$ and $\ddot{P}^d_{\theta}$.

\subsection{Short-cut Learning}

The {\em cross entropy loss} of a prediction model $\hat{P}_{\theta}(\hat{Y}|\X)$ in a source domain $P^s(\X, Y)$ is :
\begin{eqnarray*}
	\label{eq.xentropy}
	\ell_{P^s}(\hat{P}_{\theta}) = \mE_{(\x, y) \sim P^s(\X, Y)} [- \log \hat{P}_{\theta}(y|\x) ].
\end{eqnarray*}

\noindent
It is sometimes referred to as the {\em in-distribution (ID) loss}.
We can estimate the ID loss from a training set
 $\sD^s=\{\x_i, y_i\}_{i=1}^N$ from the source domain:
\begin{eqnarray*}
	\label{eq.empiricalLoss}
	\ell_{\sD^s}(\hat{P}_{\theta}) = \frac{1}{N} \sum_{i=1}^N[-\log\hat{P}_{\theta}(y_i|\x_i) ].
\end{eqnarray*}
This is known as {\em empirical cross entropy}.  The model parameters $\theta$ can be estimated by minimizing $\ell_{\sD^s}(\hat{P}_{\theta})$.  This is known as {\em empirical risk minimization (ERM)}~\cite{vapnik1998statistical}.

In the data generation model $P^s$,  $Y$ depends only on the core factors $\X^c$.
In the prediction model $\hat{P}_{\theta}$, on the other hand,
$\hat{Y}$ depends on $\X$. As $\X$  depends on both
 $\X^n$ and $\X^c$ in the data generation model,  the predicted label $\hat{Y}$  depends on them indirectly. This dependence is characterized by
$\ddot{P}_{\theta}(\hat{Y}|\X^c, \X^n)$.

ERM  is usually done using
gradient descent, which tends to latch on to strong and simple correlations between $\X$ and $Y$   \cite{pezeshki2020gradient}.
As $\X$ depends on both $\X^c$ and $\X^n$,
 gradient descent
captures not only correlations between $\X^c$ and $Y$ but also those between $\X^n$ and $Y$.
It does not distinguish between the two types of correlations, which leads
to the internalization of easy-to-learn correlations  between $\X^n$ and $Y$ into the prediction model. This  is known as {\em short-cut learning} \cite{geirhos2020shortcut}.

Short-cut learning renders the
dependence of $\hat{Y}$ on $\X^n$ in the the prediction model similar to the manner $Y$ depends on $\X^n$ in the source domain. As the dependence is spurious and is not shared by the target domain, the  quality of prediction on the target domain tends to deteriorate.  Two examples of this are shown in
Fig.~\ref{fig.spurious}.
A model that predicts wolves based on snowy background would  misclassify a husky in the snow as a wolf, and may also misclassify  wolves in other backgrounds.
Similarly, a model that predicts malignant skin lesions based on the presence of rulers would do poorly on images without rulers.

\section{Conditions for Optimal Domain Generalization}
\label{sec.conditions}

Suppose $\hat{P}_{\theta}(\hat{Y}|\X)$ is a prediction model trained on a source domain
$P^s$ from a CLD family.  In this section we establish conditions for optimal generalization to a target domain $P^t$ in the same family, i.e., the conditions for $\hat{P}_{\theta}$ to minimize the  {\em out-of-distribution (OOD)} cross entropy loss
$\ell_{P^t}(\hat{P}_{\theta})$.

 \subsection{Causal-Invariant Prediction and Causal-Faithful Prediction}
\label{sec.theo.1}

Let $\mathcal{X}^c$ and $\mathcal{X}^n$  be the sets of all possible values of
the latent variables $\X^c$ and $\X^n$ respectively.
Consider an example $\x$ generated by  $P^{*}(\X|\x^c, \x^n)$  from a pair of  values $(\x^c, \x^n) \in \mathcal{X}^c \times \mathcal{X}^n$
of  $\X^c$ and $\X^n$. Let $\tilde{\x}$ be an another example sampled from the same $\x^c$ and a different $\tilde{\x}^n$.  Formally,
 \begin{eqnarray}
 	\label{eq.2examples}
 \x \sim P^{*}(\X|\x^c, \x^n),
 \tilde{\x} \sim P^{*}(\X|\x^c, \tilde{\x}^n).
 \end{eqnarray}
The two examples ${\x}$ and  $\tilde{\x}$ contain the same semantic contents and hence should be classified into the same class.
In this sense, $\x$ and $\tilde{\x}$  make up
a  {\em contrastive pair}.
 We say that a prediction model $\hat{P}_{\theta}$ is  {\em causal-invariant} if
\begin{eqnarray}
	\label{eq.c-invariant}
	\hat{P}_{\theta}(\hat{Y}|\x) = \hat{P}_{\theta}(\hat{Y}|\tilde{\x}),
\end{eqnarray}
for any $\x$ and $\x$ sampled via (\ref{eq.2examples}) from any  $\x^c \in \mathcal{X}^c$, $\x^n \in \mathcal{X}^n$ and
$\tilde{\x}^n \in \mathcal{X}^n$.
In other words, the prediction output  does not change in response to variations in the non-core factors $\X^n$ as long as the core factors $\X^c$ remain fixed.

The concept of {\em causal-invariant prediction (CIP)}, as defined above, is not a new one and frequently appears in the DG literature in various forms \cite{mitrovic2021representation,mahajan2021domain}. It is closely related to a notion described in \cite{peters2016causal} that has a very similar name, {\em invariant causal prediction (ICP)}. One difference between the two is that we are concerned with the invariance of prediction with respect to a latent variable $\X^n$, which is entangled with another latent variable $\X^c$ to yield the observed input $\X$. Our objective is to make the prediction model invariant to $\X^n$, although we cannot observe it. In contrast, ICP is concerned with the invariance of prediction with respect to observed variables and does not address the entanglement of latent  factors.

\begin{proposition}
	\label{proposition.c-invariant}
	Let $\hat{P}_{\theta}$ be a prediction model for a CLD family and
	$\ddot{P}_{\theta}$ be the fused model defined by (\ref{eq.fused}). If  $\hat{P}_{\theta}$ is causal-invariant, then
	\begin{eqnarray}
		\ddot{P}_{\theta}(\hat{Y}|\x^c,  \x^n)=	\ddot{P}_{\theta}(\hat{Y}|\x^c, \tx^n)
		\label{eq.c-invariant2}
	\end{eqnarray}
for  any  $\x^c \in \mathcal{X}^c$, $\x^n \in \mathcal{X}^n$ and
$\tilde{\x}^n \in \mathcal{X}^n$.
\end{proposition}  	

The proof of this proposition and other proofs can be found in Appendix A.
Equation (\ref{eq.c-invariant2}) allows us to see the invariance 
of prediction $\hat{Y}$ to the non-core factors $\X^n$  more clearly.  Note that  it does not
imply  (\ref{eq.c-invariant})  in general. Equation  (\ref{eq.c-invariant}) is a stronger condition.  It means that the prediction model can filter out
the impact of   noise in example generation as well as that of the non-core factors.

 We say that a prediction model $\hat{P}_{\theta}$ is {\em causal-faithful} if
\begin{eqnarray}
	\label{eq.c-faithful}
	\hat{P}_{\theta}(\hat{Y}|\X)= P^0_{\theta}(\hat{Y}|\X^c),
\end{eqnarray}
for some conditional distribution $P^0_{\theta}(\hat{Y}|\X^c)$ of
$\hat{Y}$ given $\X^c$.
In other words, a causal-faithful model relies only on the core factors $\X^c$ for prediction.

 \noindent

 \begin{proposition}
 	\label{proposition.c-faithful}
 	In a CLD family,
 	a prediction model is causal-invariant if and only if it is
 	causal-faithful.
 \end{proposition}

 It is now clear that causal-invariant prediction (CIP)  and 
 {\em causal-faithful prediction (CFP)}  are two different perspectives on the same notion.
We will see later that they lead to different approaches to DG. The following theorem shows their implication on the cross entropy loss.

\begin{theorem}
	\label{theo.CF-loss}
	Let $\hat{P}_{\theta}(\hat{Y}=y|\X)$ be a prediction model for a CLD family, and  $P^{d}(\X^c, \X^n, \X, Y)$ be a domain from the family.
	If  the prediction model is causal-faithful, then 
	\begin{eqnarray}
		\label{eq.theo.CF-loss}
		\ell_{P^d}(\hat{P}_{\theta}) =
		\mE_{\x^c \sim P^d(\X^c)}[
		\mE_{y \sim  P^{*}(Y|\x^c)}[-\log P^0_{\theta}(\hat{Y}=y|\x^c)], ] 		
	\end{eqnarray}
	where $P^0_{\theta}(\hat{Y}=y|\x^c)$ is as in (\ref{eq.c-faithful}).
	
\end{theorem}
\noindent This theorem states that  the cross entropy loss of a causal-faithful prediction model depends only on the distribution $P^d(\X^c)$ of the core factors, and is independent of the distribution
$P^d(\X^n)$ of the non-core factors. It is intuitively straightforward.

 \subsection{Conditions for Optimal Domain Generalization}
 \label{sec.theo.2}

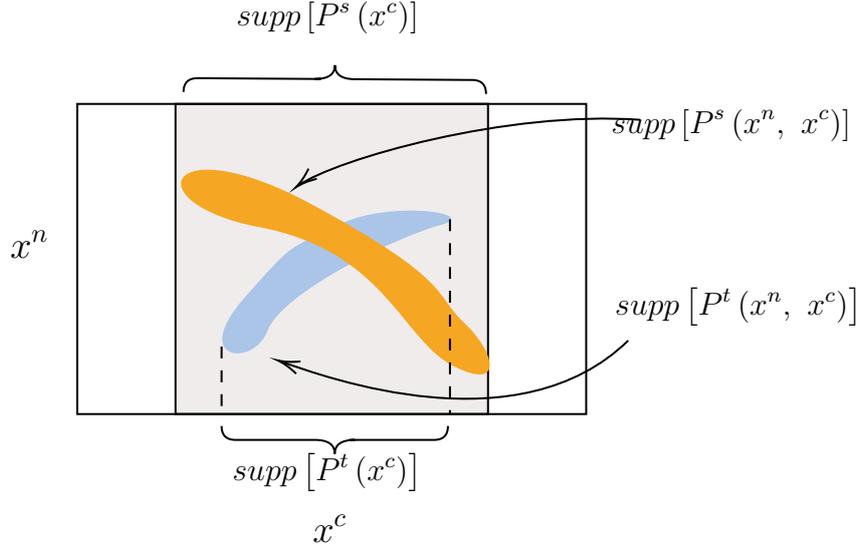
\begin{figure}[t]
	\centering
	
	\tikzset{every picture/.style={line width=0.75pt}} 

\begin{tikzpicture}[x=0.75pt,y=0.75pt,yscale=-1,xscale=1]

\draw  [fill={rgb, 255:red, 240; green, 236; blue, 236 }  ,fill opacity=1 ] (226,55) -- (382,55) -- (382,211) -- (226,211) -- cycle ;
\draw  [draw opacity=0][fill={rgb, 255:red, 74; green, 144; blue, 226 }  ,fill opacity=0.42 ] (307,116) .. controls (332,103) and (380,110) .. (358,116) .. controls (336,122) and (279,150) .. (272,169) .. controls (265,188) and (233,184) .. (260,153) .. controls (287,122) and (282,129) .. (307,116) -- cycle ;
\draw  [draw opacity=0][fill={rgb, 255:red, 245; green, 166; blue, 35 }  ,fill opacity=1 ] (268,117) .. controls (205,105) and (221,66) .. (293,105) .. controls (365,144) and (355,151) .. (370,165) .. controls (385,179) and (391,200) .. (365,187) .. controls (339,174) and (331,129) .. (268,117) -- cycle ;
\draw    (458,63) .. controls (391.7,58.13) and (306.39,82.72) .. (287.33,96) ;
\draw [shift={(286,97)}, rotate = 320.91] [color={rgb, 255:red, 0; green, 0; blue, 0 }  ][line width=0.75]    (10.93,-3.29) .. controls (6.95,-1.4) and (3.31,-0.3) .. (0,0) .. controls (3.31,0.3) and (6.95,1.4) .. (10.93,3.29)   ;
\draw    (452,174) .. controls (397.4,227.63) and (301.92,192.84) .. (278.65,184.58) ;
\draw [shift={(277,184)}, rotate = 19.29] [color={rgb, 255:red, 0; green, 0; blue, 0 }  ][line width=0.75]    (10.93,-3.29) .. controls (6.95,-1.4) and (3.31,-0.3) .. (0,0) .. controls (3.31,0.3) and (6.95,1.4) .. (10.93,3.29)   ;
\draw   (381,50) .. controls (381.03,45.33) and (378.72,42.98) .. (374.05,42.95) -- (315.55,42.57) .. controls (308.88,42.52) and (305.56,40.17) .. (305.59,35.5) .. controls (305.56,40.17) and (302.22,42.48) .. (295.55,42.43)(298.55,42.45) -- (237.05,42.05) .. controls (232.38,42.02) and (230.03,44.33) .. (230,49) ;
\draw   (177,55) -- (431,55) -- (431,211) -- (177,211) -- cycle ;
\draw   (249,217) .. controls (249,221.67) and (251.33,224) .. (256,224) -- (295.5,224) .. controls (302.17,224) and (305.5,226.33) .. (305.5,231) .. controls (305.5,226.33) and (308.83,224) .. (315.5,224)(312.5,224) -- (355,224) .. controls (359.67,224) and (362,221.67) .. (362,217) ;
\draw  [dash pattern={on 4.5pt off 4.5pt}]  (249,177) -- (249,211) ;
\draw  [dash pattern={on 4.5pt off 4.5pt}]  (363,113) -- (363,211) ;

\draw (142,117) node [anchor=north west][inner sep=0.75pt]  [font=\large] [align=left] {$\displaystyle x^{n}$};
\draw (293,260) node [anchor=north west][inner sep=0.75pt]  [font=\large] [align=left] {$\displaystyle x^{c}$};
\draw (255,2) node [anchor=north west][inner sep=0.75pt]  [font=\normalsize] [align=left] {$\displaystyle supp\left[ P^{s}\left( x^{c}\right)\right]$};
\draw (442,56) node [anchor=north west][inner sep=0.75pt]  [font=\normalsize] [align=left] {$\displaystyle supp\left[ P^{s}\left( x^{n} ,\ x^{c}\right)\right]$};
\draw (444,146) node [anchor=north west][inner sep=0.75pt]  [font=\normalsize] [align=left] {$\displaystyle supp\left[ P^{t}\left( x^{n} ,\ x^{c}\right)\right]$};
\draw (253,229) node [anchor=north west][inner sep=0.75pt]  [font=\normalsize] [align=left] {$\displaystyle supp\left[ P^{t}\left( x^{c}\right)\right]$};

\end{tikzpicture}
	
	\caption{\small Illustration of concepts and results:  A prediction model is causal-invariant if it depends only on $\X^c$ but not $\X^n$, and hence makes the same prediction on all examples sampled from the same ``vertical line".  If such a model also minimizes the cross-entropy loss of the source domain $P^s$ (i.e.,
		examples sampled from $\supp[P^s(\X^c, \X^n)]$), then it is optimal for all examples sampled from
		$\supp[P^s(\X^c)] \times \mathcal{X}^n$ (i.e., the inner rectangle), and can hence generalize optimally to any target domains
		$P^t$ with $\supp[P^t(\X^c)] \subseteq \supp[P^s(\X^c)]$.
	}
	\label{fig.theo-illustrate}
\end{figure}

\begin{theorem}
	\label{theo.ID-optimal}
	Let  $\hat{P}_{\theta}$ be a prediction model for CLD family  and $P^s$  be a source   domain  the family. Suppose
	\begin{itemize}
		\item[1).] $\hat{P}_{\theta}$ is causal-invariant/faithful, and
		\item[2).] $\hat{P}_{\theta}$ minimizes the  in-distribution (ID) loss
		$\ell_{P^s}(\hat{P}_{\theta})$.
	\end{itemize}
	\noindent
	Then, for any $\x^c \in \supp[P^s(\X^c)]$,  ${\x}^n \in \mathcal{X}^n $ and  $\x \sim P^*(\X|\x^c, \x^n)$, we have
	\begin{eqnarray}
	\label{eq.ID-optimal}
	\hat{P}_{\theta}(\hat{Y}=y|\x) = {P}^{*}(Y=y|\x^c) 
\end{eqnarray}
	for any value $y$ of $Y$ and $\hat{Y}$.

\end{theorem}
 Note that
  the requirement that
	$\ell_{P^s}(\hat{P}_{\theta})$ is minimized   means not only the  parameters
	of the prediction model  $\hat{P}_{\theta}$ is minimized, but its also architecture.   For the next theorem, we need the support
	$\supp[P^d(\X^c)] :=  \{\x^c| P^{d}(\x^c)>0\}$ of $P^d(\X^c)$. It is the set of core factors that appear in a domain $P^d$.

\begin{theorem}
	\label{theo.OOD-optimal}
 \noindent  Suppose all the conditions in Theorem \ref{theo.ID-optimal} hold.
 Let $P^t$ be a target domain from the CLD family.  If
	\[\supp[P^t(\X^c)] \subseteq \supp[P^s(\X^c)],\]
then the prediction model
$\hat{P}_{\theta}$
also minimizes the OOD loss $\ell_{P^t}(\hat{P}_{\theta})$.  In other words, it generalizes optimally to the target domain.
\end{theorem}

Fig.~\ref{fig.theo-illustrate} illustrates the notions of causal-invariant/faithful prediction and the two theorems.   Training examples $\x$ are sampled from
a latent space spanned by the values $\x^c$ and
$\x^n$ of the  latent variables $\X^c$ and
$\X^n$, which  we depict  as a two-dimensional box.  A prediction model is
causal-invariant if it makes the same prediction for examples sampled from the same ``vertical line" in the latent space.  A prediction model is
causal-faithful if it uses only $\x^c$ for prediction.

Theorem \ref{theo.ID-optimal} asserts that, if a causal-invariant/faithful prediction model $\hat{P}_{\theta}$ minimizes the cross-entropy loss of a source domain, then it makes optimal prediction not only on source domain examples (i.e., $\x$ sampled from $\supp[P^s(\X^c, \X^n)]$), but also any examples $\tilde{\x}$ sampled from $\supp[P^s(\X^c)]\times \mathcal{X}^n$.
Because $\supp[P^s(\X^c)]\times \mathcal{X}^n$ is a superset of  $\supp[P^s(\X^c, \X^n)]$,
this enables   optimal generalization
to any target domain $P^t$ such that $\supp[P^t(\X^c)] \subseteq \supp[P^s(\X^c)]$),
as stated in Theorem \ref{theo.OOD-optimal}. In summary, there are three conditions
for optimal domain generalization (ODG) in a CLD family:
\bit
\item {\em ODG Condition 1}: $\hat{P}_{\theta}$ minimizes the cross entropy loss in $P^s$,
\item {\em ODG Condition 2}: $\hat{P}_{\theta}$ is causal-invariant/faithful, 
 and
\item {\em ODG Condition 3}: All core factors present in $P^t$ also appear $P^s$, i.e.,  $\supp[P^t(\X^c)] \subseteq \supp[P^s(\X^c)]$.
\eit
\noindent We emphasize that the second condition is not only a requirement on the parameters of the prediction model, but also its capacity (architecture).
The third condition indicates that, in the CLD framework, domain generalization does not enable a prediction model to handle core factors that it has not seen during training. It can only enhance the capability of dealing to novel non-core factors and novel combinations of core and non-core factors.

Theorems \ref{theo.ID-optimal} and \ref{theo.OOD-optimal} share a similar essence with the theoretical findings of several previous works \cite{mahajan2021domain,mitrovic2021representation,wang2022out}. However, there is a crucial distinction in that the prior results are linked to particular methods for achieving the conditions of optimal domain generalization. In contrast, our theorems solely concentrate on why these conditions result in optimal domain generalization. In the following sections, we will delve into various approaches for fulfilling these conditions. By doing so, we can utilize the theorems to unify different DG approaches.

\section{Contrastive Pairs for Domain Generalization}
\label{sec.pair}

In this section, we discuss DG methods that rely on a set of contrastive pairs and training data from a single source domain. Several of these methods leverage contrastive pairs to promote causal-invariant prediction, while one method uses contrastive pairs to encourage causal-faithful prediction.

\subsection{Contrastive Pairs for Causal-Invariant Prediction}
Suppose we have a collection of contrastive pairs
$\Pi=\{(\x_k, \tilde{\x}_k)\}_{k=1}^K$. To enforce ODG Conditions 1 and 2, a simple and direct approach is to solve the following constrained optimization problem:
\begin{eqnarray*}
\min_{\theta}  &&  \ell_{P^s}(\hat{P}_{\theta}) \\
\mbox{subject to} &&  	\hat{P}_{\theta}(\hat{Y}|\X=\x_k) = \hat{P}_{\theta}(\hat{Y}|\X=\tilde{\x}_k) \hspace{0.2cm}
	\forall (\x_k, \tilde{\x}_k) \in \Pi.
\end{eqnarray*}
If we  turn the constraints into a regularization term $r_{\theta}(\Pi)$, the problem becomes:
\begin{eqnarray}
\label{eq.pair-regularization}
	\min_{\theta}  \hspace{0.2cm}  \ell_{P^s}(\hat{P}_{\theta}) + \lambda r_{\theta}(\Pi),
\end{eqnarray}

\noindent where $\lambda$ is a balancing parameter. We call this
{\em pair regularization}.  Several pair regularization methods have been proposed, which  we explain below.

Suppose  $\hat{P}_{\theta}$ consists of a feature extractor $f_{\phi}$ with parameters $\phi$  and a linear classification head $g_{\w}$ with parameters
$\w$.  Hence, $\theta = (\phi, \w)$.
For an input $\x$,  let $f_{\phi}^u(\x)$ be the component of the feature vector $f_{\phi}(\x)$ that is associated with a feature unit $u$. Let $w_{uy}$ be the
weight between the feature unit $u$ and the output unit for a class $y$.
The logit for the class $y$ is $z_{\theta}^y(\x) = \sum_u  w_{uy}f_{\phi}^u(\x)$. \footnote{Assume the bias is represented by a dummy unit.}
The regularization term can be defined in three ways:
\bit
\item {\em Probability Matching}:
$r_{\theta}(\Pi) = \mE_{(\x, \tx) \sim \Pi} [ \sD_{KL}(\hat{P}_{\theta}(\hat{Y}|\x) ||\hat{P}_{\theta}(\hat{Y}|\tx))]$
\item {\em Logit Matching}:  $r_{\theta}(\Pi) = \mE_{(\x, \tx) \sim \Pi} [ \sum_{y} (z_{\theta}^y(\x) - z_{\theta}^y(\tx))^2]$.
\item {\em Feature Matching}:
$r_{\theta}(\Pi) = \mE_{(\x, \tx) \sim \Pi} [ \sum_{u} (f_{\phi}^u(\x) - f_{\phi}^u(\tx))^2]$.
\eit
 Probability matching is utilized in {\em ReLIC} (Representation Learning via Invariant Causal Mechanisms)\cite{mitrovic2021representation}, logit matching is employed in {\em CoRE} (Conditional Variance Regularization)  \cite{heinze2021conditional}, and
 feature matching is used in {\em MatchDG} \cite{mahajan2021domain}.
 It is worth noting that while we focus on pairs for simplicity, logit and feature matching also be  applied to the case with
 groups of multiple  examples that share the same  semantic contents. To achieve this, we can simply replace the sum of squared differences with the sum of variances. This is done  in CoRE and MatchDG.

\subsection{Contrastive Pairs for Causal-Faithful Prediction}
A contrastive pair $(\x, \tx)$ share the same semantic contents. Often a label $y$ for the contents is available. In this case, we have
a {\em labeled contrastive pair} and we denote it as $(\x, \tx: y)$.

Suppose there is a collection of labeled contrastive pairs $\Pi=\{(\x_k, \tilde{\x}_k: y_k)\}_{k=1}^K$.
The following regularization term~\cite{gao2023contrastive} can be used to encourage causal-faithful prediction:
\[r_{\theta}(\Pi) = \mE_{(\x, \tx: y) \sim \Pi} [\sum_u w^2_{uy}
(f_{\phi}^{u}(\x) - f_{\phi}^u(\tx))^2] . \]

\noindent To understand the idea, imagine the ideal case where the feature extractor $f$ cleanly disentangles $\x^c$ and $\x^n$:
The feature vector $f_{\phi}(\x)$ is divided into two part $H_1(\x)$ and
$H_2(\x)$ that depend only on $\x^c$ and $\x^n$ respectively, with
$H_1(\x) = H_1(\tx)$ and $H_2(\x) \neq H_2(\tx)$. A causal-faithful predictor should predict $y$ based only on $H_1(\x)$. Hence the weights $w_{uy}$
for the feature units $u$ in $H_2(\x)$  should be 0, such that the features from the non-core factors $\x^n$ are ignored. 
In practice, it is difficult to cleanly disentangle $\x^c$ and $\x^n$.  In such a case, the regularization term has the following effects: 
\bit
\item[1).] It encourages  $g_{\w}$ to put high weights $w_{uy}$ on the units $u$ where $f_{\phi}^u(\x) \approx f_{\phi}^u(\tx)$, and 
\item[2).]  It encourages $f_{\phi}$ to make $f_{\theta}^u(\x) \approx f_{\phi}^u(\tx)$ for units $u$ with high weights $w_{uy}$.
\eit

\noindent Note that
$\sum_u w^2_{uy}
(f_{\phi}^{u}(\x) - f_{\phi}^u(\tx))^2
= \sum_u
( w_{uy}f_{\phi}^{u}(\x) -  w_{uy}f_{\phi}^u(\tx))^2$.
The regularization term  essentially matches the contributions
$w_{uy} f^u_{\phi}(\x)$ and $w_{uy} f^u_{\phi}(\tx)$ to the logit $z_{\theta}^y$ of $y$.   Hence we call it  {\em logit attribution matching (LAM)}~\cite{gao2023contrastive}.

\subsection{Creation of Contrastive Pairs}
Contrastive pairs are created in a number of ways. When analyzing the CelebA dataset~\cite{liu2015deep},  Heinze-Deml {\em et al.}~\cite{heinze2021conditional} pair up photos of the same person.  When analyzing medical images, Ouyang {\em et al.}~\cite{ouyang2022causality} create pairs by first transforming an image  using a  randomly-weighted shallow network twice and then blending
the two resulting images in a ``spatially-variable manner". The idea is to simulate different possible acquisition processes and to mitigate ``shifted-correlation effect".
 Gao {\em et al.}~\cite{gao2023contrastive} use a segmentation tool to remove the background of an image, replace it with the background of another image, and pair up the resulting image with the original one. They show that significant DG improvements can be achieved with a small number (low hundreds) of pairs.  They also show that pairs created using simple
data augmentations such as cropping, scaling, and flipping are not effective at improving DG performance.

Contrastive pairs can be learned when there are multiple source domains.
Robey {\em et al.}~\cite{robey2021model} and Wang {\em et al.}~\cite{wang2022out}  build image-to-image translation networks between different domains and use them to create pairs.
 Mahajan {\em et al.}~\cite{mahajan2021domain} propose an iterative algorithm that uses contrastive learning to map images to a latent space, and then match up images from different domains that have the same class label and
are close to each other in the latent space.

For multi-source DG problems with style shift, Gao {\em et al.}~\cite{gao2023contrastive} propose to create contrastive pairs using StableDiffusion~\cite{rombach2021highresolution}. The idea is to add a mild level of Gaussian noise to the latent
representation of an image, and then remove the noise under the
guidance of a text prompt. To create an art painting for a photo in the
PACS dataset~\cite{li2017deeper}, they use the prompt  ``a minimalist
drawing of a {\tt class\_name}, outline only, no texture".

\subsection{Discussion}

Conceptually, LAM aims to make the classification head $g$ depend only on core features. In this sense, it is a {\em $g$-oriented method}.
In contrast,  feature matching aims to make the feature representations of contrastive
pairs identical. In this sense, it is an {\em $f$-oriented method}.
Probability and logit matching aims to make the prediction outputs of the whole model  on contrastive
pairs identical.  In this sense, they are {\em $g{\circ}f$-oriented methods}.

Pair regularization is conceptually a straightforward way to achieve causal-invariant prediction, and 
 logit attribution matching is conceptually a straightforward way to achieve causal-faithful prediction.
 Their success hinges on the creation of contrastive pairs with sufficient coverage of core and non-core factors. We believe further research in this direction is warranted.

\section{Multi-Source DG Approaches}
\label{sec.multi-source}

The most popular setting for research on DG is where multiple source domains are available.  In this section we discuss multi-source DG methods in the CLD framework. We show the following statement holds under some conditions:

\begin{quote}
{\em If a prediction model for a CLD family is causal-invariant,  then some quantities/distributions/functions of the model are invariant across all domains in the family.}
\end{quote}

\noindent We present multi-source DG methods as ways to enforce the variance in those quantities/distributions/functions. 

\subsection{Variants of the CLD Model}

We need three variants of the CLD model as shown in 
Figure~\ref{fig.CLD-2}.  Different variants encapsulate different assumptions. 
The first variant {\em CLD1} 
 is the same as CLD except 
 a domain identifier $D$ is added so as to facilitate discussion of different domains.  Each value $d$ of $D$ correspond to a domain.  The joint distribution of core and non-core factors in the domain is
 $P^d(\X^c, \X^n) = P(\X^c, \X^n|D=d)$.   
 
 The second variant CLD2 assumes that $\X^c$ is independent of the domain identifier $D$, and $\X^c$ is not causally influenced by $\X^n$ or any exogenous confounding factors, although
 $\X^c$ may (or might not) causally influence $\X^n$.
 Those assumptions imply that the distribution of core factors is invariant
 across domains, i.e., 
 \begin{eqnarray}
 	\label{eq.prob-core}	
 	P^{d_1} (\X^c) = 	P^{d_2} (\X^c),
 \end{eqnarray}
 \noindent for any two domains $d_1$ and $d_2$ from a CLD2 family.  Note that covariate shift
 ($P^{d_1} (\X)  \neq 	P^{d_2} (\X)$) may still exist.

 The third variant CLD3 is  same as CLD2 except that
 $Y$ is causally influenced by $D$ and the causal direction between $\X^c$ and $Y$ reversed.  So, it is an {\em anti-causal model}. CLD3 makes weaker assumptions as compared  to  CLD2.  In particular, (\ref{eq.prob-core}) does hold in CLD3 in general. 	However, the following is true from CLD3: 
 \begin{eqnarray}
 	\label{eq.cond-prob-core}	
 	P^{d_1} (\X^c|Y) = 	P^{d_2} (\X^c|Y) = P^*(\X^c|Y).
 \end{eqnarray}
 This is a part of the causal-invariant generation assumption for the CLD3 model.

  We also need  a latent variable $\bH$  to denote the activations at the feature layer of a prediction model $\hat{P}_{\theta}(\hat{Y}|\X)$.\footnote{We refer to latent variables in the generation model as latent factors and those in the prediction model latent features.}  It is related to the input $\X$ through the feature extractor $\bH=f_{\phi}(\X)$.  Although the relationship between $\bH$ and $\X$ is deterministic, we can still think of it as a conditional distribution
$\hat{P}_{\phi}(\bH|\X)$.  The classification head $g$ defines
$\hat{P}_{\w}(\hat{Y}|\bH)$. 
As  in Section \ref{sec.fusedModel}, the prediction model can be combined with parts of the generation model to form fused models:
\begin{eqnarray*}
\ddot{P}_{\theta}(\X, \bH, \hat{Y}|\X^c, \X^n) & =&
\hat{P}_{\w}(\hat{Y}|H) \hat{P}_{\phi}(H|\X) P^*(\X|\X^c, \X^n), \\
\ddot{P}^d_{\theta}(\X, \bH, \hat{Y}, \X^c, \X^n) &=&
\ddot{P}_{\theta}(\X, \bH, \hat{Y}|\X^c, \X^n) P^d(\X^c, \X^n).
\end{eqnarray*}	
\noindent In the last model, we can talk about the marginal distribution
$\ddot{P}^d_{\theta}(H)$ of the latent features $H$ in a domain $d$  and
the conditional distribution  $\ddot{P}^d_{\theta}(H|Y)$ of the latent features $H$ given a class $Y$ in domain $d$. 
As the distributions of $H$ depends only on $\phi$, we can write them as $\ddot{P}^d_{\phi}(H)$ and $\ddot{P}^d_{\phi}(H|Y)$ respectively.

\begin{center}
	\begin{figure}[t]
		\begin{tabular}{ccc}
		\begin{tikzpicture}
			\begin{scope}[every node/.style={minimum size=2em}]
				\node (xn0) at (0, 0) {$\X^n$};
				\node (xc0) at (1.6, 0) {$\X^c$};
				\node (y)  at (3.2, 0) {$Y$};
				\node (x) at (0.8, -1.3) {$\X$};
                \node (D) at (0.8, 1.3) {$\textup{D}$};
                \node (yhat)  at (3.8, -1.3) {$\hat{Y}$};
                \node (H) at (2.4, -1.3) {$\textup{H}$};
			\end{scope}
			\begin{scope}[>={Stealth[black]},
				every edge/.style={draw=black}]
				\path [dashed] (xn0) edge node {} (xc0);
				\path [->] (xc0) edge node {} (y);
				\path [->] (xn0) edge node {} (x);
				\path [->] (xc0) edge node {} (x);
                \path [dashed] [->] (D)   edge node {} (xn0);
                \path [dashed] [->] (D)   edge node {} (xc0);
                \path [->] (x)   edge node {} (H);
                \path [->] (H)   edge node {} (yhat);
			\end{scope}
			\begin{scope}[on background layer]
				\draw[rounded corners=10, fill=black, fill opacity=0.04, draw=none]($(xn0) + (-0.6, -0.25)$) -- ($(xn0) + (-0.6, 0.35)$) -- ($(D) + (-0.3, 0.3)$)  -- ($(D) + (0.3, 0.3)$) -- ($(xc0) + (0.5, 0.4)$) -- ($(y) + (0.5, 0.4)$)  -- ($(y) + (0.5, -0.4)$) -- ($(xc0) + (0.5, -0.4)$) -- ($(x) + (0.4, -0.4)$) -- ($(x) + (-0.4, -0.4)$) -- cycle;
				\draw[rounded corners=10, fill=blue, fill opacity=0.05, draw=none]($(x) + (-0.5, -0.4)$) -- ($(x) + (-0.5, 0.4)$) -- ($(yhat) + (0.5, 0.4)$)  -- ($(yhat) + (0.5, -0.4)$) -- cycle;
			\end{scope}
		\end{tikzpicture}
		& 	\begin{tikzpicture}
			\begin{scope}[every node/.style={minimum size=2em}]
				\node (xn0) at (0, 0) {$\X^n$};
				\node (xc0) at (1.6, 0) {$\X^c$};
				\node (y)  at (3.2, 0) {$Y$};
				\node (x) at (0.8, -1.3) {$\X$};
                \node (D) at (0.8, 1.3) {$\textup{D}$};
                \node (yhat)  at (3.8, -1.3) {$\hat{Y}$};
                \node (H) at (2.4, -1.3) {$\textup{H}$};
			\end{scope}
			\begin{scope}[>={Stealth[black]},
				every edge/.style={draw=black}]
				\path [->] (xc0) edge node {} (y);
				\path [->] (xn0) edge node {} (x);
				\path [->] (xc0) edge node {} (x);
                \path [dashed] [->] (D)   edge node {} (xn0);
                \path [dashed] [->] (xc0)   edge node {} (xn0);
                \path [->] (x)   edge node {} (H);
                \path [->] (H)   edge node {} (yhat);
			\end{scope}
			\begin{scope}[on background layer]
				\draw[rounded corners=10, fill=black, fill opacity=0.04, draw=none]($(xn0) + (-0.6, -0.25)$) -- ($(xn0) + (-0.6, 0.35)$) -- ($(D) + (-0.3, 0.3)$)  -- ($(D) + (0.3, 0.3)$) -- ($(xc0) + (0.5, 0.4)$) -- ($(y) + (0.5, 0.4)$)  -- ($(y) + (0.5, -0.4)$) -- ($(xc0) + (0.5, -0.4)$) -- ($(x) + (0.4, -0.4)$) -- ($(x) + (-0.4, -0.4)$) -- cycle;
				\draw[rounded corners=10, fill=blue, fill opacity=0.05, draw=none]($(x) + (-0.5, -0.4)$) -- ($(x) + (-0.5, 0.4)$) -- ($(yhat) + (0.5, 0.4)$)  -- ($(yhat) + (0.5, -0.4)$) -- cycle;
			\end{scope}
		\end{tikzpicture}
		& 	\begin{tikzpicture}
			\begin{scope}[every node/.style={minimum size=2em}]
				\node (xn0) at (0, 0) {$\X^n$};
				\node (xc0) at (1.6, 0) {$\X^c$};
				\node (y)  at (3.2, 0) {$Y$};
				\node (x) at (0.8, -1.3) {$\X$};
                \node (D) at (0.8, 1.3) {$\textup{D}$};
                \node (yhat)  at (3.8, -1.3) {$\hat{Y}$};
                \node (H) at (2.4, -1.3) {$\textup{H}$};
                \path [dashed] [->] (xc0)   edge node {} (xn0);
			\end{scope}
			\begin{scope}[>={Stealth[black]},
				every edge/.style={draw=black}]
				\path [->] (y) edge node {} (xc0);
				\path [->] (xn0) edge node {} (x);
				\path [->] (xc0) edge node {} (x);
                \path [dashed] [->] (D)   edge node {} (xn0);
                \path [->] (x)   edge node {} (H);
                \path [->] (H)   edge node {} (yhat);
                \path [dashed] [->] (D)   edge node {} (y);
			\end{scope}
			\begin{scope}[on background layer]
				\draw[rounded corners=10, fill=black, fill opacity=0.04, draw=none]($(xn0) + (-0.6, -0.25)$) -- ($(xn0) + (-0.6, 0.35)$) -- ($(D) + (-0.3, 0.3)$)  -- ($(D) + (0.3, 0.3)$) -- ($(y) + (0.5, 0.4)$)  -- ($(y) + (0.5, -0.4)$) -- ($(xc0) + (0.5, -0.4)$) -- ($(x) + (0.4, -0.4)$) -- ($(x) + (-0.4, -0.4)$) -- cycle;
				\draw[rounded corners=10, fill=blue, fill opacity=0.05, draw=none]($(x) + (-0.5, -0.4)$) -- ($(x) + (-0.5, 0.4)$) -- ($(yhat) + (0.5, 0.4)$)  -- ($(yhat) + (0.5, -0.4)$) -- cycle;
			\end{scope}
		\end{tikzpicture} \\
		\scriptsize (a) CLD1  & \scriptsize  (b) CLD2  & \scriptsize (c) CLD3
	\end{tabular}
		\caption{\small Variants of the CLD model: (a)  CLD1 is the same
		as the CLD model except with a domain identifier $D$ added to the generation model, (b) CLD2
	is the same as CLD1 except that $\X^c$ is assumed independent of $D$, (c) CLD3 is the same as CLD2 except the causal direction between $\X^c$ and $Y$ is reversed, and $Y$ causally depends on $D$. In all the three variants, $\bH$ is the output of a hidden layer in the prediction model.}
		\label{fig.CLD-2}
	\end{figure}
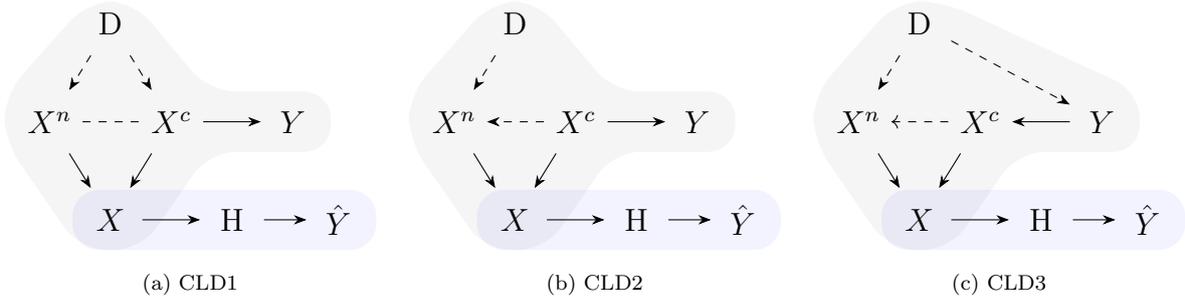
\end{center}

\subsection{Risk Matching}
\label{sec.multi-risk}

\begin{proposition}
	\label{pro.riskMatching}
	Let $\hat{P}_{\theta}$ be a prediction model for a CLD2 family and
	$P^{d_1}$ and $P^{d_2}$ be any two domains from the family.  If $\hat{P}_{\theta}$ is causal-invariant, then
	\begin{eqnarray}
	\label{eq.riskMatching}	
		\ell_{P^{d_1}} (\hat{P}_{\theta}) = 	\ell_{P^{d_2}} (\hat{P}_{\theta}).
	\end{eqnarray}
\end{proposition}

 This proposition states  that, in the CLD2 family,  
  causal-invariant prediction   implies that the cross entropy  loss is invariant across all domains in the family.  Conversely, matching  the losses in different domains would encourage a prediction model to be causal-invariant and 
 hence help with domain generalization. This turns out to be the key idea of a DG algorithm known as
{\em V-REx} (variance risk extrapolation)~\cite{krueger2020outofdistribution}.
Suppose there are $K$ training domains $\{d_1, \ldots, d_k\}$.
V-REx uses the variance of the losses in the $K$ domains as a regularization term:
\begin{eqnarray*}
	\min_{\theta}  \hspace{0.2cm}
	\mbox{V-REx}(\theta)  :=
	\frac{1}{K}\sum_{d=1}^K \ell_{P^{d}}(\hat{P}_{\theta}) + \lambda \mbox{Var}(\{\ell_{P^{d_1}}(\hat{P}_{\theta}), \ldots, \ell_{P^{d_K}}(\hat{P}_{\theta})\}).
\end{eqnarray*}
\noindent The hyperparameter  $\lambda \in [0, 1)$ controls the balance between reducing average loss and enforcing equality of losses, with $\lambda = 0$
recovering ERM, and $\lambda \rightarrow \infty$  leading V-REx to focus
entirely on making the losses equal. 
\footnote{Note that, historically,
V-REx originated from an effort to improve another algorithm known as  {\em Group DRO} (group distributionally robust optimization)~\cite{sagawa2019distributionally}, which minimizes the worst-case loss:
\begin{eqnarray*}
	\min_{\theta}  \hspace{0.2cm}
	\mbox{DRO}(\theta)  :=
	\max_{k=1}^K  \hspace{0.1cm} \ell_{P^{d_k}}(\hat{P}_{\theta}).
\end{eqnarray*}
}

\subsection{Gradient Matching}
\label{sec.multi-gradient}

We say that a prediction model $\hat{P}_{\theta}$ is {\em strongly causal-invariant}
if it is causal-invariant and $\hat{P}_{\theta'}$  is causal-invariant for all
$\theta'$ in a small neighborhood of $\theta$. 
\begin{proposition}
	\label{pro.gradientMatching}
Let $\hat{P}_{\theta}$ be a prediction model for a CLD2 family and
$P^{d_1}$ and $P^{d_2}$ be any two domains from the family. If $\hat{P}_{\theta}$ is strongly causal-invariant,
	\begin{eqnarray}
		\label{eq.gradientMatching}	
		\nabla_{\theta}\ell_{P^{d_1}} (\hat{P}_{\theta}) = 	\nabla_{\theta}	\ell_{P^{d_2}} (\hat{P}_{\theta}).
	\end{eqnarray}
\end{proposition}

This proposition states that
strongly causal-invariant prediction implies that 
 the gradients of the cross entropy  loss are invariant across all domains in 
 a  CLD2 family.  
 Conversely, matching the gradients in different domains would
encourage a prediction model to be  strongly causal-invariant and hence help with domain generalization.~\footnote{Intuitively, gradient matching may be more effective when used in combination with risk matching.}  Several DG methods are based on this idea, including {\em Fish}~\cite{shi2021gradient}, {\em IGA} (inter gradient alignment)~\cite{koyama2020out},  {\em AND-mask}~\cite{parascandolo2021learning}, and {\em Fishr}~\cite{rame2021fishr}.

Here is the objective function used in  Fish:
\begin{eqnarray*}
	\min_{\theta}  \hspace{0.2cm}
	\mbox{Fish}(\theta)  :=
	\frac{1}{K}\sum_{k=1}^K \ell_{P^{d_k}}(\hat{P}_{\theta}) - \lambda \frac{1}{K(K-1)} \sum_{d \neq d' \in \{d_1, \ldots, d_K\}}
	G_{d}(\theta) \cdot G_{d'}(\theta),
\end{eqnarray*}
\noindent where $G_d(\theta)$ is the gradient vector 	$\nabla_{\theta}\ell_{P^{d}} (\hat{P}_{\theta})$ in domain $d$, and 	$G_{d}(\theta) \cdot G_{d'}(\theta)$ is the  inner product of two gradient vectors. The regularization term encourages the inner product to be large and thereby aligns the gradients of different domains.
 IGA aims to achieve the same goal using the following objective function:
\begin{eqnarray*}
	\min_{\theta}  \hspace{0.2cm}
	\mbox{IGA}(\theta)  :=
	\frac{1}{K}\sum_{k=1}^K \ell_{P^{d_k}}(\hat{P}_{\theta}) + \lambda \hspace{0.1cm}  \mbox{trace} ( \mbox{Var}(\{G_{d_1}(\theta), \ldots, G_{d_K}(\theta)\})),
\end{eqnarray*}
\noindent where Var$(...)$ is the empirical covariance matrix of the $K$ gradient vectors. Minimizing the trace of this matrix pushes each component of $\ell_{P^{d_k}}(\hat{P}_{\theta})$ toward the same across domains.
 AND-mask simply zeros out gradient components that have inconsistent signs across domains at each iteration of training.  Specifically, any  component of a gradient vector in a domain whose sign is not shared the majority of other domains is set to 0.

The gradient vector
	$\nabla_{\theta}\ell_{P^{d}}$ that we have been discussing is the average
	of per-example gradient vectors of  training examples in domain $d$.
{Fishr computes  the {\em domain-level variance of per-example gradients} for each domain and  minimizes the variance among those domain-level gradient variances across domains. The motivation is that ``the domain-level
gradients should be similar, not only in average direction,
but also (and maybe mostly importantly) in statistics such as variance and disagreements".

 {\em MLDG}
(meta-learning for domain generalization)~\cite{li2018learning} divides
training domains into meta-train and meta-test domains. Each step of  its training process is designed to improve the model not only  in the meta-train domains but also  in the meta-test domains.  An analysis of the algorithm reveals that it essentially maximizes the inner product of the gradients in the meta-train and meta-test domains besides minimizing the loss in the meta-train and meta-test domains.  In other words, it is similar to Fish conceptually.

\subsection{Feature Distribution Matching}
\label{sec.multi-representation}

We say that   
 a feature extractor $f_{\phi}(\X)$ is {\em causal-invariant} if
\begin{eqnarray}
	\label{eq.H-c-invariant}
	f_{\phi}(\x) = f_{\phi}(\tx)
\end{eqnarray}
for any contrastive pair $(\x, \tx)$.
Similar to Proposition
\ref{proposition.c-faithful}, we can show that $f_{\phi}$ is
causal-invariant if and only if  it is  {\em causal-faithful} in the 
sense that
\begin{eqnarray}
	\label{eq.H-c-faithful}
	f_{\phi}(\X) = f^0_{\phi}(\X^c) 
\end{eqnarray}
for some function $f^0_{\phi}(\X^c)$ of $\X^c$.~\footnote{We only need to, in the proofs, replace $\hat{Y}$ with $H$ and observe that deterministic relationship is a special case of probabilistic relationship.}
 The features $H$ extracted by such an extractor are {\em causal-invariant/faithful features}.  It is evident that
 a prediction model $\hat{P}_{\theta}$ is causal-invariant/faithful if its feature extractor is.

\begin{proposition}
	\label{prop.IRL}
Let $\hat{P}_{\theta}$ be a prediction model for a CLD2 family and
$P^{d_1}$ and $P^{d_2}$ be any two domains from the family. If  the feature
extractor $f_{\phi}(\X)$ of  $\hat{P}_{\theta}$ is causal-invariant, then
	\begin{eqnarray}
		\label{eq.IRL}
		\ddot{P}^{d_1}_{\phi}(\bH) = 	\ddot{P}^{d_2}_{\phi}(\bH).
	\end{eqnarray}

\end{proposition}

Equation (\ref{eq.IRL}) means that the distributions of data points from different domains  in the latent space are identical. This property is known as {\em domain-invariant representations}. It is not to be confused with the notion of causal-invariant features, which  refers to the invariance of the latent representation of a particular data point to changes in non-core factors change.
 Proposition \ref{prop.IRL} establishes that, in the CLD2 family,
 causal-invariant features  imply domain-invariant representations.  Conversely, learning domain-invariant representations should help make the feature extractor causal-invariant, and hence improve domain generalization.

In fact, {\em domain-invariant representation learning} is a popular approach to
 multi-source DG.   Representative methods include  {\em DANN} (domain-adversarial neural network)~\cite{ganin2016domain}, {\em CORAL} (correlation
 alignment)~\cite{sun2016deep}, and {\em MMD-AAE} (maximum mean
 discrepancy-based adversarial autoencoder)~\cite{li2018domain}.  DANN builds two classifiers on top of the latent features $\bH$, one for class labels and another for domain labels.  The loss of the label classifier is minimized so as to make $\bH$ predictive of the classes. On the other hand,  the loss of the domain classifier is maximized so as to make $\bH$ unpredictive (and hence independent) of domains $D$.  As such, DANN is a {\em domain adversarial learning} method.

   In contrast, CORAL and MMD-AAE are {\em domain alignment} methods.
 CORAL  aligns the distributions
 $\ddot{P}_{\theta}^d(\bH)$ for different domains by matching their second-order statistics, while
 MMD-AAE aligns the distributions by minimizing the maximum mean discrepancy between them.

\subsection{Conditional Feature Distribution Matching}
\label{sec.C-DANN}

{\em
C-DANN} (conditional domain adversarial neural networks)~\cite{li2018deep} is  another popular domain-invariant representation learning method. To understand C-DANN, consider the model CLD3 shown in Figure~\ref{fig.CLD-2} (c).  
\begin{proposition}
	\label{prop.IRL2}
		Let $\hat{P}_{\theta}$ be a prediction model for a CLD3 family and 
		$P^{d_1}$ and $P^{d_2}$ be any two domains from the  family. If  the feature
		extractor $f_{\phi}(\X)$ of  $\hat{P}_{\theta}$ is causal-invariant, then
	\begin{eqnarray}
		\label{eq.IRL2}
		\ddot{P}^{d_1}_{\phi}(\bH|Y) = 	\ddot{P}^{d_2}_{\phi}(\bH|Y).
	\end{eqnarray}
     Consequently,
     \begin{eqnarray}
     	\label{eq.IRL3}
     	\bar{{P}}^{d_1}_{\phi}(\bH|) = \bar{{P}}^{d_2}_{\phi}(\bH),
     \end{eqnarray}
 where $\bar{{P}}^d_{\phi}(\bH) := \sum_{y=1}^{|Y|}\ddot{P}^d_{\phi}(\bH|Y=y) \frac{1}{|Y|}$ and $|Y|$ in turn is the number of classes.
\end{proposition}

Equation (\ref{eq.IRL2}) means that the  {\em class-conditional distributions}
$\ddot{P}^{d}_{\phi}(\bH|Y)$ of data points from different domains in the latent space are identical.
Equation (\ref{eq.IRL3}) means that the same is true for the   {\em class prior-normalized marginals}
 $\bar{{P}}^d_{\theta}(\bH)$.
C-DANN builds $|Y|+2$ classifiers on top of the latent features $\bH$,  one for class labels and $|Y|+1$ for  domain labels.   There is one class-conditional domain classifier for each class to enforce equation (\ref{eq.IRL2}), and there is an additional domain classifier to enforce equation (\ref{eq.IRL3}).   The loss of the label classifier is minimized, while the losses of the domain classifiers are maximized.

 An advantage of C-DANN over DANN is that it does not assume $P^d(\X^c)$ is invariant across domains. However, there is a theoretical caveat.  In the CLD3 model, $P^d(Y|\X^c)$ may change across domains.  If a causal-invariant model
 $\hat{P}_{\theta}(\hat{Y}|X)$ minimizes the ID loss of a source domain, then
  $\hat{P}_{\theta}(\hat{Y}|X) =P^s(Y|\X^c)$. Consequently, it cannot be optimal for any target domain $P^t$ with  $P^t(Y|\X^c) \neq P^s(Y|\X^c)$.
In other words,   optimal domain generalization is not possible if $P^d(Y|\X^c)$ changes across domains.

\subsection{Invariant Risk Minimization }
\label{sec.multi-IRM}
A CLD2 model is said to be {\em deterministic}
 if $\X$ depends on $\X^c$ and $\X^n$ deterministically, i.e.,
$\X = f^*(\X^c, \X^n)$.
Let $\hat{P}_{\theta}$ be a prediction model for a deterministic CLD2
family with feature extractor $\bH=f_{\phi}(\X)$ and classification head $\hat{P}_{\w}(\hat{Y}|\bH)$.  As the parameters  are divided into two
parts $(\phi, \w)$,  we write  $\ell_{P^d}(\hat{P}_{\theta})$ as $\ell_{P^d}(\phi, \w)$ for simplicity.
Define
$\ddot{f}_{\phi}(\X^c, \X^n) = f_{\phi}(f^*(\X^c, \X^n))$. If the feature extractor is  causal-faithful,  then $\ddot{f}_{\phi}(\X^c, \X^n) = f^0_{\phi}(\X^c)$.

\begin{proposition}
	\label{pro.IRM}
	Let  $\hat{P}_{\theta}$ be a prediction model  for the a deterministic CLD2 family
	and let $P^d$ be any domain from the family.  
	Further let $\w^d_{\phi} = \arg \min 	\ell_{P^d}(\phi, \w)$.  If the feature extractor $f_{\phi}(\X)$ of $\hat{P}_{\theta}$ is causal-faithful, then $\w^d_{\phi}$ is invariant to $d$.
\end{proposition}

The proposition states that, for a fixed (not necessarily optimal)
causal-faithful feature extractor,
the losses of all domains in a deterministic CLD2 family are minimized at the same value of $\w$.   Conversely, enforcing the losses of all source domains be minimized at the same value of $\w$ should encourage the feature extractor to be causal-faithful, which in turn helps with domain generalization.  This leads to the following objective that {\em IRM} (invariant risk minimization)~\cite{arjovsky2020out} optimizes:
	\begin{eqnarray}
		\label{eq.IRM}
		min_{\phi, \w}	& & \frac{1}{K} \sum_{k=1}^K \ell_{P^{d_k}}(\phi, \w)\\
		\mbox{subject to}		&&    \w \in \arg \min   \ell_{P^{d_k}}(\phi, \w),
		\forall k \in \{1, \ldots, K\}. \nonumber
	\end{eqnarray}

\subsection{Discussion }
\label{sec.multi-source-discuss}

IRM aims to make the classification head $g_{\w}$ optimal for all domains
and thereby making the prediction model causal-faithful. It is therefore a $g$-oriented method.  The feature distribution matching algorithms are $f$-oriented methods, while the risk and gradient matching algorithms are $g{\circ}f$-oriented methods.  They aims to make the prediction model causal-invariant.

The single-source+pairs algorithms discussed in the previous section directly target at making a prediction model causal-invariant/faithful.  In contrast, the multi-source algorithms discussed in this section hope to achieve the same goal indirectly by enforcing constraints implied by causal-invariant/faithful prediction.  In other words, they aim to achieve some necessary (but not sufficient) conditions  for causal-invariant/faithful prediction.
Additionally, they require more assumptions, as encapsulated in the CLD2 and CLD3 models.   The reason is that, as compared with multiple training domains,  contrastive pairs contain stronger information that can help a classifier focus on the core factors.  On the other hand, it is generally believed to be easier to obtain multiple training domains than to create a large number of contrastive pairs.

\section{Single-Source DG Approaches}
\label{sec.single-source}


Domain generalization with a single source domain and no auxiliary information is a challenging task. However, there are several methods that have been proposed to improve model performance in target domains in this setting.  In this section we discuss  those methods in the CLD framework.

\subsection{Approximately Causal-Invariant Prediction}
\label{sec.pcip}

A key condition for optimal domain generalization is that a prediction model needs to be causal-invariant (ODG Condition 2).  This condition implies that equation (\ref{eq.c-invariant2}) must hold.  In practice we can only make the equation hold approximately and hence can only achieve {\em approximately causal-invariant prediction (PCIP)}.  To quantify the degree to which a prediction  model $\hat{P}_{\theta}(\hat{Y}|\x)$ satisify the condition, we define
the {\em CI index} as measured in a domain $d$ to be:
\begin{eqnarray}
	\label{eq.dg-index}
	\mbox{CI-Index}^d(\theta) := 1-\mE_{(\x^c, \x^n) \sim P^d(\X^c, \X^n), \tx^n \sim
	P^d(\X^n) }[JSD(\ddot{P}_{\theta}(\hat{Y}|\x^c, \x^n)|| \ddot{P}_{\theta}(\hat{Y}|\x^c, \tx^n))],
\end{eqnarray}
\noindent where JSD is the Jensen-Shannon divergence with logarithm base 2. It is 1 when  equation (\ref{eq.c-invariant2}) holds exactly, and it is 0
when $\supp[\ddot{P}_{\theta}(\hat{Y}|\x^c, \x^n)] \cap \supp[\ddot{P}_{\theta}(\hat{Y}|\x^c, \tx^n)] = \emptyset$ for all
$\x^c$, $\x^n$ and $\tx^n$.

There are ways to increase the CI index in the single-source setting.
One idea is to suppress  strong  dependencies of $\hat{Y}$ on both $\X^n$ and $\X^c$.~\footnote{It would be ideal to reduce the dependence of $\hat{Y}$ on  $\X^n$ only, but this is difficult because $\X^c$ and $\X^n$ are entangled in the training example.}  Two single-source DG algorithms can be viewed as instantiations of this idea. 
{\em SD} (spectral decoupling)~\cite{pezeshki2021gradient} suppress  strong  dependencies of $\hat{Y}$ on both $\X^n$ and $\X^c$  by regularizing the logits:

\begin{eqnarray*}
	\min_{\theta}  \hspace{0.2cm}
	\mbox{SD}(\theta)
	 :=\mE_{(\x, y) \sim P^s(\X, Y)} [- \log \hat{P}_{\theta}(\hat{Y}=y|\x) + \lambda
	 ||\z_{\theta}(\x)||^2],
\end{eqnarray*}
\noindent where $\z_{\theta}(\x)$ is the logit vector of the input $\x$.
A component of the logit vectors typically depends strongly on a subset of the features $H$.  If it is reduced, the dependence of $\hat{Y}$ on those features are reduced as a consequence. 
{\em RSC} (representation self-challenging)~\cite{huang2020self}  aims to achieve the same goal in a heuristic manner. 
  At each iteration of training, it mutes  the feature units associated with the
highest gradients, such that the network is forced to predict the labels through other less salient
features.

We have  estimated the CI indices of
three models trained on the same dataset by ERM, SD and RSC.  The details are reported in Appendix B.  The results are given in the following table.    We see that SD is indeed effective in increasing the CI index.

\begin{center}
\begin{table}[h!]
\centering
\begin{tabular}{c|l}
			\hline
			Methods & CI index \\ \hline
			ERM & 0.5428\\ \hline
   			RSC &  0.5412 \\ \hline
			SD &  0.6619 \\ \hline
\end{tabular}
\label{tab:CI Index}
\end{table}
\end{center}

\subsection{Data-Oriented Methods}
\label{sec.data-oriented}
While causal-invariant prediction is important for optimal domain generalization in general, good generalization to target domains may sometimes be achieved without it.

\begin{theorem}
	\label{theo.odg-no-ci}
	Let $\hat{P}_{\theta}$ be a prediction model for a CLD family and 
	 $P^s$  be a source   domain from the family. 
 If $\hat{P}_{\theta}$ minimizes the  ID loss
		$\ell_{P^s}(\hat{P}_{\theta})$,
	then 	
	
	\bit
	\item[1).]  For any $(\x^c, \x^n) \in \supp[P^s(\X^c, \X^n)]$,
	$\x \sim P^*(\X|\x^c, x^n)$ and $y \sim P^*(Y|\x^c)$, we have
	\begin{eqnarray}
		\label{eq.odg-no-ci}
	\hat{P}_{\theta}(\hat{Y}=y|\x) = {P}^{*}(Y=y|\x^c).
\end{eqnarray}
	
	\item[2).] For any target domain $P^t$ from the family such that $$\supp[P^t(\X^c, \X^n)] \subseteq \supp[P^s(\X^c, \X^n)],$$ $\hat{P}_{\theta}$ minimizes the OOD loss
		$\ell_{P^t}(\hat{P}_{\theta}).$
	\eit
\end{theorem}

 Note that the first part of
 Theorem \ref{theo.odg-no-ci} is the same as the conclusion of Theorem \ref{theo.ID-optimal}, except
 for the scope where the equality (\ref{eq.ID-optimal})/(\ref{eq.odg-no-ci}) holds.
 In Theorem \ref{theo.OOD-optimal}, the conditions  are stronger and   the equality holds for any $(\x^c, \x^n) \in \supp[P^s(\X^c)] \times \mathcal{X}^n$, which is a superset of  $\supp[P^s(\X^c, \X^n)]$   (see Fig.~\ref{fig.theo-illustrate} to visualize the two sets).

According to Theorem \ref{theo.odg-no-ci}, optimal model performance in the target domain  can be achieved  with ODG Condition 1 and the following condition:

\bit
\item {\em ODG Condition 3'}: All core factors, non-core factors, and their combinations that are  present in $P^t$ also appear $P^s$, i.e.,  $\supp[P^t(\X^c, \X^n)] \subseteq \supp[P^s(\X^c, \X^n)]$.
\eit
 
 \noindent ODG Condition 3' is a very strong condition. Because of this, Theorem \ref{theo.odg-no-ci} is hardly a result on domain generalization. All core and non-core factors in the target domain are seen in the source domain, although they might follow a different distribution.  However, the theorem suggests one approach to increase model performance in target domains, i.e., to increase the coverage of core and non-core factors in the source domain.  This can be done     by either augmenting existing data or collecting a large amount data.~\footnote{
 	In practice, the ID loss is approximated by the empirical loss (\ref{eq.empiricalLoss}).  The use of more data also helps to make
 	the empirical loss a better approximation of the theoretical ID loss.}

 Data augmentation involves creating more data by modifying existing ones.
 One simple method to do this is {\em Mixup}~\cite{zhang2017mixup}. It creates new training examples by interpolating existing examples and their labels, and is often used as a baseline for DG.  Other methods include data randomization, adversarial augmentation, and the use of generative models~\cite{wang2022generalizing}.
 {\em SagNet} (Style-Agnostic Network)~\cite{nam2019reducing}
 generates additional data online by swapping the style of an input with the style of another using AdaIN~\cite{huang2017arbitrary}.

 Large datasets cover a wide range of core and non-core factors, as well as  their combinations, resulting large $\supp[P^s(\X^c, \X^n)]$.  
 This broad coverage makes models trained on large datasets more likely to generalize well, particularly in the case of zero-shot learning, where fine-tuning is not required. This observation has been made in various studies, including~\cite{radford2021learning}.   However, fine-tuning is often necessary and it can  lead to significant drop in domain generalization performance~\cite{kumar2022fine,radford2021learning}.
The reason is that fine-tuning reduces the scope of core and non-core factors that the model can handle.

\subsection{Feature Disentanglement Methods}
{\em Feature disentanglement methods} aim to separate core and non-core factors during training, which can help it generalize to new domains that contain different combinations of these factors. To understand them from the perspective of the CLD framework, note that $\supp[P^s(\X^c)P^s(\X^n)] = \supp[P^s(\X^c)] \times \supp[P^s(\X^n)]$
is a superset of $[\supp[P^s(\X^c, \X^n)]$   (see Fig.~\ref{fig.theo-illustrate} to visualize the two sets).   Consequently, the estimation of
$P^s(\X^c)P^s(\X^n)$ increases the coverage of the combinations of core and non-core factors. 
This is the idea behind feature disentanglement, which has been explored in several DG methods, including {\em StableNet}~\cite{kuang2020stable}, {\em CSG} (causal semantic generative model)~\cite{liu2021learning}, and {\em NuRD} (nuisance-randomized distillation)~\cite{puli2021out}.

\subsection{Ensemble Learning Methods}
\label{sec.ensemble}

Several ensemble learning methods have recently  been proposed for domain generalization, 
 including  {\em SWAD} (stochastic weight
averaging densely)~\cite{cha2021swad}, {\em EoA} (ensembling moving average)~\cite{arpit2022ensemble}, {\em MIRO} (mutual information regularization with oracle)~\cite{cha2022domain}, and {\em SIMPLE} (model-sample matching for domain generalization)~\cite{li2022simple}.  In experiments with 283 pretrained models~\cite{li2022simple}, those methods outperformed common DG algorithms on several DG benchmarks~\cite{li2022simple}, especially when  models pretrained on large datasets were included in the ensemble.

From the perspective of the CLD framework, ensemble learning can help with domain generalization for two reasons. First, Theorem \ref{theo.odg-no-ci} requires that the architecture and parameters of a prediction model be optimal for a source domain, which is seldom satisfied by a model with a specific architecture and trained in a specific way. Using an ensemble of pretrained models with different architectures and training procedures can help alleviate this problem. Second, the pretrained models may have been trained on different datasets, which exposes the ensembled model to more core and non-core factors than any of the original models and helps with domain generalization according to the second part of Theorem \ref{theo.odg-no-ci}.     
However, it should be noted that relying on a large number of pre-trained models might be problematic in real-world applications.

\subsection{Discussion}

Domain generalization is both an academic problem and a practical problem.
As an academic problem, 
it aims to enable a prediction model to handle two types of novelty in a target domain: (1)  non-core factors that were unseen in the source domain, and (2) combinations of core and non-core factors that were unseen in the source domain. PCIP methods can help address both types of novelty by reducing the reliance of a prediction model on salient non-core factors. Feature disentanglement methods can help address the second type of novelty by considering all combinations of core and non-core factors during training.

As a practical problem, domain generalization aims to prevent deterioration of model performance in a target domain.  Data-oriented methods help with this by increasing the coverage of core factors, non-factors, and their combinations in the training data. Ensemble 
 methods help also mainly because they improve model performance in the source domain, which is observed to be  linearly correlated with model performance in the target domain~\cite{taori2020measuring,miller2021accuracy}.

\section{A Summary of DG Methods}
\label{sec.summary}

\input{summary_figure.tex}

We have discussed a host of DG methods in the CLD framework.  Fig.~\ref{fig.summary} provides a summary of the discussions.  There are two routes to make
a prediction model $\hat{P}_{\theta}$, trained in a source domain $P^s$,  generalize optimally to a target domain $P^t$:

\begin{itemize}
	\item[1).]  Make $\hat{P}_{\theta}$ causal-invariant/faithful (ODG Condition 2) and ensure $\supp[P^t(\X^c)] \subseteq \supp[P^s(\X^c)]$ (ODG Condition 3), or
	\item[2).]  Ensure $\supp[P^t(\X^c, \X^n)] \subseteq \supp[P^s(\X^c, X^n)]$  (ODG Condition 3').
\end{itemize}
\noindent  In both routes, the cross entropy loss in the source domain needs to be minimized (ODG Condition 1).
 The data-oriented and feature disentanglement methods follow the second route.  The ensemble methods follow neither route and focus only on ODG Condition 1.

Most DG methods follow the first route, where the key is to make a prediction model causal-variant/faithful. In the case of a single source domain, we can make a prediction model approximately causal-invariant by reducing the reliance on salient feature as in RSC and  SD.  When contrastive pairs are available, we can directly apply  CIP constraints  to the feature extractor $f$ as in MatchDG, or apply CIP constraints to the whole model $g{\circ}f$ as in ReLIC and CoRE. We can also apply CFP constraints to $g$ as in LAM.  

In the case of multiple source domains, CIP implies invariance in feature distributions, losses and gradients.  Such implied invariance can be used to constrain the feature extractor $f$ as in CORAL, MMD and DANN, or to constrain the whole model $g{\circ}f$  as in V-Rex, Fish, Fishr, IGA, MLDG and AND-Mask.   CFP implies simultaneous optimality of the classification head across domains, which is exploited in IRM.



\section{Conclusion}
\label{sec.conclusion}

In this paper, we have presented a novel causal framework for domain generalization, which builds upon notions and ideas from previous literature. Our framework emphasizes the separation of the data generation model and the prediction model, and the use of a fused model to bridge them. We have shown that this approach allows for the establishment of optimal conditions for domain generalization in a mathematically rigorous and intuitively meaningful way.

Furthermore, our framework also emphasizes the separation of the conditions for optimal domain generalization and the methods aimed at achieving those conditions. This provides a common foundation for understanding different domain generalization approaches, which in turn enables insights into their relative strengths and weaknesses.

Our work provides a clear and concise summary of the state-of-the-art in domain generalization research, and we hope it inspires new work in this critical area of machine learning. The proposed framework can serve as a starting point for future research and can help guide the development of new approaches for overcoming the challenges posed by domain shift.

\subsection*{Acknowledgment}
Research on this paper was supported by  Hong Kong Research Grants Council under grant 16204920,  and Huawei Technologies Co., Ltd. under Project
19201840HWLB15Z015 and Grant YBN2020045058. Kaican Li and Weiyan Xie were supported in part by the Huawei PhD Fellowship Scheme. 
{\small

\bibliography{main}

\begin{thebibliography}{10}

\bibitem{ahuja2021invariance}
Kartik Ahuja, Ethan Caballero, Dinghuai Zhang, Jean-Christophe Gagnon-Audet,
  Yoshua Bengio, Ioannis Mitliagkas, and Irina Rish.
\newblock Invariance principle meets information bottleneck for
  out-of-distribution generalization.
\newblock {\em Advances in Neural Information Processing Systems},
  34:3438--3450, 2021.

\bibitem{alcorn2019strike}
Michael~A Alcorn, Qi~Li, Zhitao Gong, Chengfei Wang, Long Mai, Wei-Shinn Ku,
  and Anh Nguyen.
\newblock Strike (with) a pose: Neural networks are easily fooled by strange
  poses of familiar objects.
\newblock In {\em CVPR}, 2019.

\bibitem{arjovsky2020out}
Martin Arjovsky.
\newblock {\em Out of distribution generalization in machine learning}.
\newblock PhD thesis, New York University, 2020.

\bibitem{arjovsky2019invariant}
Martin Arjovsky, L{\'e}on Bottou, Ishaan Gulrajani, and David Lopez-Paz.
\newblock Invariant risk minimization.
\newblock {\em arXiv:1907.02893}, 2019.

\bibitem{arpit2022ensemble}
Devansh Arpit, Huan Wang, Yingbo Zhou, and Caiming Xiong.
\newblock Ensemble of averages: Improving model selection and boosting
  performance in domain generalization.
\newblock {\em Advances in Neural Information Processing Systems},
  35:8265--8277, 2022.

\bibitem{beery2018recognition}
Sara Beery, Grant~Van Horn, and Pietro Perona.
\newblock Recognition in terra incognita.
\newblock In {\em ECCV}, 2018.

\bibitem{cha2021swad}
Junbum Cha, Sanghyuk Chun, Kyungjae Lee, Han-Cheol Cho, Seunghyun Park, Yunsung
  Lee, and Sungrae Park.
\newblock Swad: Domain generalization by seeking flat minima.
\newblock {\em Advances in Neural Information Processing Systems},
  34:22405--22418, 2021.

\bibitem{cha2022domain}
Junbum Cha, Kyungjae Lee, Sungrae Park, and Sanghyuk Chun.
\newblock Domain generalization by mutual-information regularization with
  pre-trained models.
\newblock In {\em European Conference on Computer Vision}, pages 440--457.
  Springer, 2022.

\bibitem{degrave2021ai}
Alex~J DeGrave, Joseph~D Janizek, and Su-In Lee.
\newblock Ai for radiographic covid-19 detection selects shortcuts over signal.
\newblock {\em Nature Machine Intelligence}, 3(7):610--619, 2021.

\bibitem{deng2009imagenet}
Jia Deng, Wei Dong, Richard Socher, Li-Jia Li, Kai Li, and Li~Fei-Fei.
\newblock Imagenet: A large-scale hierarchical image database.
\newblock In {\em CVPR}, 2009.

\bibitem{dosovitskiy2020image}
Alexey Dosovitskiy, Lucas Beyer, Alexander Kolesnikov, Dirk Weissenborn,
  Xiaohua Zhai, Thomas Unterthiner, Mostafa Dehghani, Matthias Minderer, Georg
  Heigold, Sylvain Gelly, et~al.
\newblock An image is worth 16x16 words: Transformers for image recognition at
  scale.
\newblock In {\em ICLR}, 2020.

\bibitem{ganin2016domain}
Yaroslav Ganin, Evgeniya Ustinova, Hana Ajakan, Pascal Germain, Hugo
  Larochelle, Fran{\c{c}}ois Laviolette, Mario Marchand, and Victor Lempitsky.
\newblock Domain-adversarial training of neural networks.
\newblock {\em JMLR}, 2016.

\bibitem{gao2023contrastive}
Han Gao, Kaican Li, Yongxiang Huang, Luning Wang, Chen~Caleb Cao, and
  Nevin~Lianwen Zhang.
\newblock Contrastive domain generalization via logit attribution matching.
\newblock {\em under submission}, 2023.

\bibitem{geirhos2020shortcut}
Robert Geirhos, Jörn-Henrik Jacobsen, Claudio Michaelis, Richard Zemel,
  Wieland Brendel, Matthias Bethge, and Felix~A. Wichmann.
\newblock Shortcut learning in deep neural networks.
\newblock {\em Nature Machine Intelligence}, 2020.

\bibitem{heinze2021conditional}
Christina Heinze-Deml and Nicolai Meinshausen.
\newblock Conditional variance penalties and domain shift robustness.
\newblock {\em Machine Learning}, 110(2):303--348, 2021.

\bibitem{hoover1990logic}
Kevin~D Hoover.
\newblock The logic of causal inference: Econometrics and the conditional
  analysis of causation.
\newblock {\em Economics \& Philosophy}, 6(2):207--234, 1990.

\bibitem{huang2017arbitrary}
Xun Huang and Serge Belongie.
\newblock Arbitrary style transfer in real-time with adaptive instance
  normalization.
\newblock In {\em ICCV}, 2017.

\bibitem{huang2020self}
Zeyi Huang, Haohan Wang, Eric~P Xing, and Dong Huang.
\newblock Self-challenging improves cross-domain generalization.
\newblock {\em arXiv:2007.02454}, 2020.

\bibitem{koyama2020out}
Masanori Koyama and Shoichiro Yamaguchi.
\newblock Out-of-distribution generalization with maximal invariant predictor.
\newblock {\em arXiv:2008.01883}, 2020.

\bibitem{krueger2020outofdistribution}
David Krueger, Ethan Caballero, Joern-Henrik Jacobsen, Amy Zhang, Jonathan
  Binas, Remi~Le Priol, and Aaron Courville.
\newblock Out-of-distribution generalization via risk extrapolation (rex).
\newblock {\em arXiv:2003.00688}, 2020.

\bibitem{kuang2020stable}
Kun Kuang, Ruoxuan Xiong, Peng Cui, Susan Athey, and Bo~Li.
\newblock Stable prediction with model misspecification and agnostic
  distribution shift.
\newblock In {\em AAAI}, 2020.

\bibitem{kumar2022fine}
Ananya Kumar, Aditi Raghunathan, Robbie Jones, Tengyu Ma, and Percy Liang.
\newblock Fine-tuning can distort pretrained features and underperform
  out-of-distribution.
\newblock {\em arXiv preprint arXiv:2202.10054}, 2022.

\bibitem{li2017deeper}
D.~Li, Y.~Yang, Y.~Song, and T.~M. Hospedales.
\newblock Deeper, broader and artier domain generalization.
\newblock In {\em ICCV}, 2017.

\bibitem{li2018learning}
Da~Li, Yongxin Yang, Yi-Zhe Song, and Timothy Hospedales.
\newblock Learning to generalize: Meta-learning for domain generalization.
\newblock In {\em AAAI}, 2018.

\bibitem{li2018domain}
Haoliang Li, Sinno Jialin~Pan, Shiqi Wang, and Alex~C Kot.
\newblock Domain generalization with adversarial feature learning.
\newblock In {\em CVPR}, 2018.

\bibitem{li2018deep}
Ya~Li, Xinmei Tian, Mingming Gong, Yajing Liu, Tongliang Liu, Kun Zhang, and
  Dacheng Tao.
\newblock Deep domain generalization via conditional invariant adversarial
  networks.
\newblock In {\em Proceedings of the European Conference on Computer Vision
  (ECCV)}, pages 624--639, 2018.

\bibitem{li2022simple}
Ziyue Li, Kan Ren, Xinyang Jiang, Yifei Shen, Haipeng Zhang, and Dongsheng Li.
\newblock Simple: Specialized model-sample matching for domain generalization.
\newblock In {\em The Eleventh International Conference on Learning
  Representations}, 2022.

\bibitem{liu2021learning}
Chang Liu, Xinwei Sun, Jindong Wang, Haoyue Tang, Tao Li, Tao Qin, Wei Chen,
  and Tie-Yan Liu.
\newblock Learning causal semantic representation for out-of-distribution
  prediction.
\newblock In {\em NeurIPS}, 2021.

\bibitem{liu2021heterogeneous}
Jiashuo Liu, Zheyuan Hu, Peng Cui, Bo~Li, and Zheyan Shen.
\newblock Heterogeneous risk minimization.
\newblock In {\em International Conference on Machine Learning}, pages
  6804--6814. PMLR, 2021.

\bibitem{liu2015deep}
Ziwei Liu, Ping Luo, Xiaogang Wang, and Xiaoou Tang.
\newblock Deep learning face attributes in the wild.
\newblock In {\em Proceedings of the IEEE international conference on computer
  vision}, pages 3730--3738, 2015.

\bibitem{lv2022causality}
Fangrui Lv, Jian Liang, Shuang Li, Bin Zang, Chi~Harold Liu, Ziteng Wang, and
  Di~Liu.
\newblock Causality inspired representation learning for domain generalization.
\newblock In {\em Proceedings of the IEEE/CVF Conference on Computer Vision and
  Pattern Recognition}, pages 8046--8056, 2022.

\bibitem{mahajan2021domain}
Divyat Mahajan, Shruti Tople, and Amit Sharma.
\newblock Domain generalization using causal matching.
\newblock In {\em International Conference on Machine Learning}, pages
  7313--7324. PMLR, 2021.

\bibitem{miller2021accuracy}
John~P Miller, Rohan Taori, Aditi Raghunathan, Shiori Sagawa, Pang~Wei Koh,
  Vaishaal Shankar, Percy Liang, Yair Carmon, and Ludwig Schmidt.
\newblock Accuracy on the line: on the strong correlation between
  out-of-distribution and in-distribution generalization.
\newblock In {\em International Conference on Machine Learning}, pages
  7721--7735. PMLR, 2021.

\bibitem{mitrovic2021representation}
Jovana Mitrovic, Brian McWilliams, Jacob~C Walker, Lars~Holger Buesing, and
  Charles Blundell.
\newblock Representation learning via invariant causal mechanisms.
\newblock In {\em ICLR}, 2021.

\bibitem{nam2019reducing}
Hyeonseob Nam, HyunJae Lee, Jongchan Park, Wonjun Yoon, and Donggeun Yoo.
\newblock Reducing domain gap via style-agnostic networks.
\newblock {\em arXiv:1910.11645}, 2019.

\bibitem{narla2018automated}
Akhila Narla, Brett Kuprel, Kavita Sarin, Roberto Novoa, and Justin Ko.
\newblock Automated classification of skin lesions: from pixels to practice.
\newblock {\em Journal of Investigative Dermatology}, 138(10):2108--2110, 2018.

\bibitem{ouyang2022causality}
Cheng Ouyang, Chen Chen, Surui Li, Zeju Li, Chen Qin, Wenjia Bai, and Daniel
  Rueckert.
\newblock Causality-inspired single-source domain generalization for medical
  image segmentation.
\newblock {\em IEEE Transactions on Medical Imaging}, 2022.

\bibitem{parascandolo2021learning}
Giambattista Parascandolo, Alexander Neitz, ANTONIO ORVIETO, Luigi Gresele, and
  Bernhard Sch{\"o}lkopf.
\newblock Learning explanations that are hard to vary.
\newblock In {\em ICLR}, 2021.

\bibitem{peters2016causal}
Jonas Peters, Peter B{\"u}hlmann, and Nicolai Meinshausen.
\newblock Causal inference by using invariant prediction: identification and
  confidence intervals.
\newblock {\em Journal of the Royal Statistical Society: Series B (Statistical
  Methodology)}, 78(5):947--1012, 2016.

\bibitem{pezeshki2021gradient}
Mohammad Pezeshki, Oumar Kaba, Yoshua Bengio, Aaron~C Courville, Doina Precup,
  and Guillaume Lajoie.
\newblock Gradient starvation: A learning proclivity in neural networks.
\newblock {\em Advances in Neural Information Processing Systems},
  34:1256--1272, 2021.

\bibitem{pezeshki2020gradient}
Mohammad Pezeshki, S{\'e}kou-Oumar Kaba, Yoshua Bengio, Aaron Courville, Doina
  Precup, and Guillaume Lajoie.
\newblock Gradient starvation: A learning proclivity in neural networks.
\newblock {\em arXiv:2011.09468}, 2020.

\bibitem{puli2021out}
Aahlad Puli, Lily~H Zhang, Eric~K Oermann, and Rajesh Ranganath.
\newblock Out-of-distribution generalization in the presence of
  nuisance-induced spurious correlations.
\newblock {\em arXiv preprint arXiv:2107.00520}, 2021.

\bibitem{radford2021learning}
Alec Radford, Jong~Wook Kim, Chris Hallacy, Aditya Ramesh, Gabriel Goh,
  Sandhini Agarwal, Girish Sastry, Amanda Askell, Pamela Mishkin, Jack Clark,
  et~al.
\newblock Learning transferable visual models from natural language
  supervision.
\newblock In {\em International Conference on Machine Learning}, pages
  8748--8763. PMLR, 2021.

\bibitem{rame2021fishr}
Alexandre Rame, Corentin Dancette, and Matthieu Cord.
\newblock Fishr: Invariant gradient variances for out-of-distribution
  generalization.
\newblock {\em arXiv:2109.02934}, 2021.

\bibitem{recht2019imagenet}
Benjamin Recht, Rebecca Roelofs, Ludwig Schmidt, and Vaishaal Shankar.
\newblock Do imagenet classifiers generalize to imagenet?
\newblock In {\em ICML}, 2019.

\bibitem{ribeiro2016should}
Marco~Tulio Ribeiro, Sameer Singh, and Carlos Guestrin.
\newblock " why should i trust you?" explaining the predictions of any
  classifier.
\newblock In {\em Proceedings of the 22nd ACM SIGKDD international conference
  on knowledge discovery and data mining}, pages 1135--1144, 2016.

\bibitem{robey2021model}
Alexander Robey, George Pappas, and Hamed Hassani.
\newblock Model-based domain generalization.
\newblock In {\em NeurIPS}, 2021.

\bibitem{rombach2021highresolution}
Robin Rombach, Andreas Blattmann, Dominik Lorenz, Patrick Esser, and Björn
  Ommer.
\newblock High-resolution image synthesis with latent diffusion models, 2021.

\bibitem{sagawa2019distributionally}
Shiori Sagawa, Pang~Wei Koh, Tatsunori~B Hashimoto, and Percy Liang.
\newblock Distributionally robust neural networks for group shifts: On the
  importance of regularization for worst-case generalization.
\newblock {\em arXiv preprint arXiv:1911.08731}, 2019.

\bibitem{shen2021towards}
Zheyan Shen, Jiashuo Liu, Yue He, Xingxuan Zhang, Renzhe Xu, Han Yu, and Peng
  Cui.
\newblock Towards out-of-distribution generalization: A survey.
\newblock {\em arXiv preprint arXiv:2108.13624}, 2021.

\bibitem{sheth2022domain}
Paras Sheth, Raha Moraffah, K~Sel{\c{c}}uk Candan, Adrienne Raglin, and Huan
  Liu.
\newblock Domain generalization--a causal perspective.
\newblock {\em arXiv preprint arXiv:2209.15177}, 2022.

\bibitem{shi2021gradient}
Yuge Shi, Jeffrey Seely, Philip~HS Torr, N~Siddharth, Awni Hannun, Nicolas
  Usunier, and Gabriel Synnaeve.
\newblock Gradient matching for domain generalization.
\newblock {\em arXiv:2104.09937}, 2021.

\bibitem{sun2016deep}
Baochen Sun and Kate Saenko.
\newblock Deep coral: Correlation alignment for deep domain adaptation.
\newblock In {\em ECCV}, 2016.

\bibitem{taori2020measuring}
Rohan Taori, Achal Dave, Vaishaal Shankar, Nicholas Carlini, Benjamin Recht,
  and Ludwig Schmidt.
\newblock Measuring robustness to natural distribution shifts in image
  classification.
\newblock {\em Advances in Neural Information Processing Systems},
  33:18583--18599, 2020.

\bibitem{tenenbaum1996separating}
Joshua Tenenbaum and William Freeman.
\newblock Separating style and content.
\newblock {\em Advances in neural information processing systems}, 9, 1996.

\bibitem{vapnik1998statistical}
Vladimir Vapnik.
\newblock {\em Statistical Learning Theory}.
\newblock Wiley, 1998.

\bibitem{wang2022generalizing}
Jindong Wang, Cuiling Lan, Chang Liu, Yidong Ouyang, Tao Qin, Wang Lu, Yiqiang
  Chen, Wenjun Zeng, and Philip Yu.
\newblock Generalizing to unseen domains: A survey on domain generalization.
\newblock {\em IEEE Transactions on Knowledge and Data Engineering}, 2022.

\bibitem{wang2022out}
Ruoyu Wang, Mingyang Yi, Zhitang Chen, and Shengyu Zhu.
\newblock Out-of-distribution generalization with causal invariant
  transformations.
\newblock In {\em Proceedings of the IEEE/CVF Conference on Computer Vision and
  Pattern Recognition}, pages 375--385, 2022.

\bibitem{xiao2020noise}
Kai Xiao, Logan Engstrom, Andrew Ilyas, and Aleksander Madry.
\newblock Noise or signal: The role of image backgrounds in object recognition.
\newblock {\em arXiv preprint arXiv:2006.09994}, 2020.

\bibitem{zech2018variable}
John~R Zech, Marcus~A Badgeley, Manway Liu, Anthony~B Costa, Joseph~J Titano,
  and Eric~Karl Oermann.
\newblock Variable generalization performance of a deep learning model to
  detect pneumonia in chest radiographs: a cross-sectional study.
\newblock {\em PLoS medicine}, 15(11):e1002683, 2018.

\bibitem{zhang2017mixup}
Hongyi Zhang, Moustapha Cisse, Yann~N Dauphin, and David Lopez-Paz.
\newblock mixup: Beyond empirical risk minimization.
\newblock {\em arXiv preprint arXiv:1710.09412}, 2017.

\bibitem{zhang2013domain}
Kun Zhang, Bernhard Sch{\"o}lkopf, Krikamol Muandet, and Zhikun Wang.
\newblock Domain adaptation under target and conditional shift.
\newblock In {\em International conference on machine learning}, pages
  819--827. PMLR, 2013.

\bibitem{zhang2021deep}
Xingxuan Zhang, Peng Cui, Renzhe Xu, Linjun Zhou, Yue He, and Zheyan Shen.
\newblock Deep stable learning for out-of-distribution generalization, 2021.

\bibitem{zhou2021domain}
Kaiyang Zhou, Ziwei Liu, Yu~Qiao, Tao Xiang, and Chen~Change Loy.
\newblock Domain generalization in vision: A survey.
\newblock {\em arXiv preprint arXiv:2103.02503}, 2021.

\end{thebibliography}
}


\newpage

\section*{Appendix A:  Proofs}

\noindent {\bf Proof of Proposition \ref{proposition.c-invariant}}:
\begin{eqnarray*}
\ddot{P}_{\theta}(\hat{Y}|\x^c,  \x^n) &=&
\mE_{\x \sim P^{*}(\X|\x^c, \x^n) }  [\hat{P}_{\theta}(\hat{Y}|\x)]
\hspace{2.2cm}
\mbox{($\because$ (\ref{eq.fused}))}  \nonumber \\
&=&
\mE_{\tx \sim P^{*}(\X|\x^c, \tx^n) }  [\hat{P}_{\theta}(\hat{Y}|\tx)] 
\hspace{2.2cm}
\mbox{($\because$ (\ref{eq.c-invariant}))}  \nonumber \\
&=& 	\ddot{P}_{\theta}(\hat{Y}|\x^c,  \tx^n). \qed
\end{eqnarray*}

\noindent {\bf Proof of Proposition \ref{proposition.c-faithful}}:
The if-part is trivially true. For the  only-if part, we need to show how the distribution $P^0_{\theta}(\hat{Y}|\X^c)$ in (\ref{eq.c-faithful})  can be constructed given (\ref{eq.c-invariant}) --- the condition for causal-invariant prediction.

Let $d_0$ be a domain  such that
$\supp[P^{d_0}(\X^c, \X^n)] = \mathcal{X}^c \times \mathcal{X}^n$.
We combine $P^{d_0}(\X^c, \X^n)$ with the fused model $\ddot{P}_{\theta}(\X, \hat{Y}|\X^c, \X^n)$ to get
\begin{eqnarray*}
\label{eq.doubleFused}
\ddot{P}^{d_0}_{\theta}(\X, \hat{Y}, \X^c, \X^n)  = \ddot{P}_{\theta}(\X, \hat{Y}|\X^c, \X^n)P^{d_0}(\X^c, \X^n).
\end{eqnarray*}
Now we can talk about
$\ddot{P}^{d_0}_{\theta}(\hat{Y}|\X^c)$, and we have
\begin{eqnarray}
\label{eq.p-d0}
\ddot{P}^{d_0}_{\theta}(\hat{Y}|\X^c) =
\int \ddot{P}_{\theta}(\hat{Y}|\X^c, \X^n)P^{d_0}(\X^n|\X^c)
\diff \X^n.
\end{eqnarray}
\noindent  Essentially we integrate $\X^n$ from $\ddot{P}_{\theta}(\hat{Y}|\X^c, \X^n)$ using $P^{d_0}(\X^n|\X^c)$.  For a given value $\x^c$ of $\X^c$, the right hand side can be written as  $\mE_{\x^n \sim P^{d_0}(\X^n|\x^c)}[
\ddot{P}_{\theta}(\hat{Y}|\x^c, \x^n)]$.

Next we will show that
(\ref{eq.c-faithful})  is satisfied with $P^0_{\theta}(\hat{Y}|\X^c)=\ddot{P}^{d_0}_{\theta}(\hat{Y}|\X^c)$.   Let $\x \sim P^*(\X|\x^c, \x^n)$ for a fixed pair  $(\x^c, \x^n) \in \mathcal{X}^c \times \mathcal{X}^n$. We have
\begin{eqnarray*}
\hat{P}_{\theta}(\hat{Y}|\x) &=&
\mE_{\tx \sim P^*(\X|\x^c, \tx^n), \tx^n \sim P^{d_0}(\X^n|\x^c)}
[\hat{P}_{\theta}(\hat{Y}|\x)]  \hspace{0.5cm}
\mbox{(expectation of constant)}\\
&=&
\mE_{\tx \sim P^*(\X|\x^c, \tx^n), \tx^n \sim P^{d_0}(\X^n|\x^c)}
[\hat{P}_{\theta}(\hat{Y}|\tx)]  \hspace{0.5cm}
\mbox{($\because$ (\ref{eq.c-invariant}))}\\
&=& 	\mE_{\tx^n \sim P^{d_0}(\X^n|\x^c)}[
\mE_{\tx \sim P^*(\X|\x^c, \tx^n)}
[\hat{P}_{\theta}(\hat{Y}|\tx)] ]\\
&=& 	\mE_{\tx^n \sim P^{d_0}(\X^n|\x^c)}[
\ddot{P}_{\theta}(\hat{Y}|\x^c, \tx^n)]  \hspace{2.0cm}
\mbox{($\because$ (\ref{eq.fused}))}\\
&=& \ddot{P}^{d_0}_{\theta}(\hat{Y}|\x^c).
\end{eqnarray*}
The only-if part is proved. $\qed$
\\

\noindent
{\bf Proof of Theorem 	\ref{theo.CF-loss}}:  
\begin{eqnarray*}
	\ell_{P^d}(\hat{P}_{\theta})  &=&\mE_{(\x, y) \sim P^d(\X, Y)} [- \log \hat{P}_{\theta}(\hat{Y}=y|\x) ]  \\
	&=&\mE_{(\x^c, \x^n) \sim P^d(\X^c, \X^n), \x \sim P^*(\X|\x^c, \x^n), y \sim P^*(Y|\x^c)} [ -\log \hat{P}_{\theta}(\hat{Y}=y|\x) ]  \\
	&=&\mE_{(\x^c, \x^n) \sim P^d(\X^c, \X^n), \x \sim P^*(\X|\x^c, \x^n), y \sim P^*(Y|\x^c)} [- \log P^0_{\theta} (\hat{Y}=y|\x^c) ]  \hspace{0.5cm}
	\mbox{($\because$ (\ref{eq.c-faithful}))} \\
	&=&\mE_{\x^c \sim P^d(\X^c), y \sim P^*(Y|\x^c)} [- \log P_0(\hat{Y}=y|\x^c) ]  \\
	&=&	\mE_{\x^c \sim P^d(\X^c)}[
	\mE_{y \sim  P^{*}(Y|\x^c)}[-\log P^0_{\theta}(\hat{Y}=y|\x^c)]].   \qed 		
\end{eqnarray*}

\noindent
{\bf Proof of Theorem 	\ref{theo.ID-optimal}}: By Theorem \ref{theo.CF-loss}, we have
\begin{eqnarray*}
	\ell_{P^s}(\hat{P}_{\theta})  = 	- \mE_{\x^c \sim P^s(\X^c)}[
	\mE_{y \sim  P^{*}(Y|\x^c)}[\log P^0_{\theta}(\hat{Y}=y|\x^c)].
\end{eqnarray*}

\noindent
As the ID $\ell_{P^s}(\hat{P}_{\theta})$ loss is minimized (over all model parameters and all model architectures),
the inner integration is maximized for any $\x^c$ such that  $P^s(\x^c)>0$.  By Gibbs' inequality, this implies that
$$\ddot{P}^{0}_{\theta}(\hat{Y}=y|\x^c)=P^*({Y}=y|\x^c)$$
for any    value $y$ of $Y$ and $\hat{Y}$.  By  (\ref{eq.c-faithful}), we get
$$\hat{P}_{\theta}(\hat{Y}=y|\x)=P^*({Y}=y|\x^c).$$
The theorem follows. $\qed$\\

\noindent
{\bf Proof of Theorem 	\ref{theo.OOD-optimal}}: By Theorem \ref{theo.CF-loss}, we have
\begin{eqnarray*}
	\ell_{P^t}(\hat{P}_{\theta}) = - \mE_{\x^c \sim P^t(\X^c)}[
	\mE_{y \sim  P^{*}(Y|\x^c)}[\log P^0_{\theta}(\hat{Y}=y|\x^c)].
\end{eqnarray*}

\noindent We know from the proof of Theorem \ref{theo.ID-optimal} that the inner expectation is maximized for all $\x^c$ such that $P^s(\x^c)>0$. Because $\supp[P^t(\X^c)] \subseteq \supp[P^s(\X^c)]$, it is also maximized for any
$\x^c$ such that $P^t(\x^c)>0$. Therefore, the OOD loss $\ell_{P^t}(\hat{P}_{\theta})$
is minimized. 
$\qed$ \\

\noindent
{\bf Proof of Proposition	\ref{pro.riskMatching}}:
We know from 
Theorem \ref{theo.CF-loss} that 
	\begin{eqnarray*}
	\ell_{P^d}(\hat{P}_{\theta}) =
	\mE_{\x^c \sim P^d(\X^c)}[
	\mE_{y \sim  P^{*}(Y|\x^c)}[-\log P^0_{\theta}(\hat{Y}=y|\x^c)], ] 		
\end{eqnarray*}
for any domain $P^d$ from a CLD family.  As  CLD2 is  a specialization of CLD, the equation holds also any domain $P^d$ from a CLD2 family.
 We also know that equation (\ref{eq.prob-core}) holds for any two domains $d_1$ and $d_2$ from a CLD2 family. The proposition follows.  $\qed$ \\

\noindent
{\bf Proof of Proposition	\ref{pro.gradientMatching}}:
This proposition follows readily from Proposition 
\ref{pro.riskMatching} and strong causal-invariant prediction.  $\qed$ \\

\noindent {\bf Proof of Proposition \ref{prop.IRL}}:   Another way to write equation (\ref{eq.H-c-faithful}) is:
\[\hat{P}_{\phi}(H|\X) = P^0_{\phi}(H|\X^c),\] 
where $P^0_{\phi}(H|\X^c)$ is the conditional distribution determined by the function
$H = f^0_{\phi}(\X^c)$.  For any domain $d$ from a CLD2 family, we have
\begin{eqnarray*}
	\ddot{P}^d_{\phi}(\bH)
	&=&  \mE_{(\x^c, \x^n) \sim P^d(\X^c, \X^n), \x \sim P^*(\X|\x^c, \x^n) }[\ddot{P}^d_{\phi}(\bH|\x)]  \\
	&=&  \mE_{(\x^c, \x^n) \sim P^d(\X^c, \X^n), \x \sim P^*(\X|\x^c, \x^n) }[\hat{P}_{\phi}(\bH|\x)]  \\
	&=&  \mE_{(\x^c, \x^n) \sim P^d(\X^c, \X^n), \x \sim P^*(\X|\x^c, \x^n) }[P^0_{\phi}(\bH|\x^c)] \\
	&=&  \mE_{\x^c \sim P^d(\X^c)}[P^0_{\phi}(\bH|\x^c)].
\end{eqnarray*}
Together with (\ref{eq.prob-core}), this implies
$\ddot{P}^{d_1}_{\phi}(\bH) =\ddot{P}^{d_2}_{\phi}(\bH)$ for any two domains $d_1$ and $d_2$ from the  CLD2 family. 
The proposition follows.  $\qed$ \\

\noindent {\bf Proof of Proposition \ref{prop.IRL2}}:   
For any domain $d$ from a CLD3 family, we have 
\begin{eqnarray*}
	\ddot{P}^d_{\phi}(\bH|Y)
		&=&  \mE_{(\x^c, \x^n) \sim P^d(\X^c, \X^n|Y), \x \sim P^*(\X|\x^c, \x^n) }[\ddot{P}^d_{\phi}(\bH|\x)]  \\
	&=&  \mE_{(\x^c, \x^n) \sim P^d(\X^c, \X^n|Y), \x \sim P^*(\X|\x^c, \x^n) }[\hat{P}_{\phi}(\bH|\x)]  \\
	&=&  \mE_{(\x^c, \x^n) \sim P^d(\X^c, \X^n|Y), \x \sim P^*(\X|\x^c, \x^n) }[P^0_{\phi}(\bH|\x^c)] \\
	&=&  \mE_{\x^c \sim P^d(\X^c|Y)}[P^0_{\phi}(\bH|\x^c)],
\end{eqnarray*}
\noindent  
where the third equation holds because the feature extractor is causal-faithful
and $P^0_{\phi}(\bH|\x^c)$ is the conditional distribution that corresponds to  the function $H=f^0_{\phi}(\x^c)$. 
Together with (\ref{eq.cond-prob-core}), this implies
$\ddot{P}^{d_1}_{\phi}(\bH|Y) =\ddot{P}^{d_2}_{\phi}(\bH|Y)$ for any two domains $d_1$ and $d_2$ from the  CLD3 family.  $\qed$ \\

\noindent
{\bf Proof of Proposition	\ref{pro.IRM}}:   
For any domain $P^d$ from a deterministic CLD2 family, we have
\begin{eqnarray*}
	\ell_{P^d}(\phi, \w)  &=&\mE_{(\x, y) \sim P^d(\X, Y)} [- \log \hat{P}_{\theta}(\hat{Y}=y|\x) ]  \\
	&=&\mE_{(\x^c, \x^n) \sim P^d(\X^c, \X^n), \x \sim P^*(\X|\x^c, \x^n), y \sim P^*(Y|\x^c)} [ -\log \hat{P}_{\theta}(\hat{Y}=y|\x) ]  \\
	&=&\mE_{(\x^c, \x^n) \sim P^d(\X^c, \X^n),  y \sim P^*(Y|\x^c)} [- \log \hat{P}_{\w}(\hat{Y}=y|\ddot{f}_{\phi}(\x^c, \x^n)) ]  \\
	& =&
	\mE_{(\x^c, \x^n) \sim P^d(\X^c, \X^n)}[ \mE_{y \sim P^*(Y|\x^c)} [- \log \hat{P}_{\w}(\hat{Y}=y|\ddot{f}_{\phi}(\x^c, \x^n)) ]].
\end{eqnarray*}
When the feature extractor is causal-faithful, we have
\begin{eqnarray*}\ell_{P^d}(\phi, \w) &=&
	\mE_{(\x^c, \x^n) \sim P^d(\X^c, \X^n)}[ \mE_{y \sim P^*(Y|\x^c)} [- \log \hat{P}_{\w}(\hat{Y}=y|f^0_{\phi}(\x^c) ]]\\
	&=&
	\mE_{\x^c \sim P^d(\X^c)}[ \mE_{y \sim P^*(Y|\x^c)} [- \log \hat{P}_{\w}(\hat{Y}=y|f^0_{\phi}(\x^c) ]].
\end{eqnarray*}
Together with (\ref{eq.prob-core}), this implies that $\ell_{P^{d_1}}(\phi, \w)
	= \ell_{P^{d_2}}(\phi, \w)$ for any two domains $d_1$ and $d_2$ from the deterministic CLD2 family.  The proposition follows. $\qed$  \\ 

\noindent
{\bf Proof of Theorem 	\ref{theo.odg-no-ci}}: For any domain $P^d$ from the CLD family,  we have
\begin{eqnarray}
	\ell_{P^d}(\hat{P}_{\theta})  &=&\mE_{(\x, y) \sim P^d(\X, Y)} [- \log \hat{P}_{\theta}(\hat{Y}=y|\x) ]  \nonumber \\
	&=&\mE_{(\x^c, \x^n) \sim P^d(\x^c, \x^n), \x \sim P^*(\X|\x^c, \x^n), y \sim P^*(Y|\x^c)} [- \log \hat{P}_{\theta}(\hat{Y}=y|\x) ]  \nonumber \\
	&=&\mE_{(\x^c, \x^n) \sim P^d(\x^c, \x^n), \x \sim P^*(\X|\x^c, \x^n)}[ \mE_{y \sim P^*(Y|\x^c)} [- \log \hat{P}_{\theta}(\hat{Y}=y|\x) ]].      \label{eq.odg-no-ci-1}		
\end{eqnarray}
\noindent  Note that those equations hold for any prediction models, not only the
causal-invariant ones as required in Theorem \ref{theo.CF-loss}.

It is given that the ID loss  $\ell_{P^s}(\hat{P}_{\theta})$ is minimized.  By Gibbs' inequality, (\ref{eq.odg-no-ci-1}) as applied to $P^s$ implies that (\ref{eq.odg-no-ci}) holds
for any $(\x^c, \x^n) \in \supp[P^s(\X^c, \X^n)]$,
$\x \sim P^*(\X|\x^c, x^n)$ and $y \sim P^*(Y|\x^c)$.  The first part is proved.

To prove the  second part, note that the condition $\supp[P^t(\X^c, \X^n)] \subseteq \supp[P^s(\X^c, \X^n)]$ implies
(\ref{eq.odg-no-ci}) holds
for any $(\x^c, \x^n) \in \supp[P^t(\X^c, \X^n)]$,
$\x \sim P^*(\X|\x^c, x^n)$ and $y \sim P^*(Y|\x^c)$.
Together with  (\ref{eq.odg-no-ci-1}), this implies that 
$\ell_{P^t}(\hat{P}_{\theta})$ is minimized.
$\qed$   \\

\section*{Appendix B:  Estimation of CI Index}
The CI Index has been defined in (\ref{eq.dg-index}), which measures the extent of a prediction model satisfying the {\em CIP} condition. Here we introduce the details of its estimation.

\subsection*{Appendix B.1: Dataset}
\textbf{ImageNet-9}~\cite{xiao2020noise} is a dataset with background shift across three domains: \emph{original}, \emph{only-fg}, and \emph{mixed-rand}.
The first domain consists of original images from ImageNet~\cite{deng2009imagenet}, while the other two domains are derived from the first domain, one with pure-black backgrounds (only-fg) and the other with random backgrounds (mixed-rand).
In our experiments, we use the original domain as the training domain and mixed-rand as the test domain.

\subsection*{Appendix B.2: Methods}
Here we take baseline method \textbf{ERM}~\cite{vapnik1998statistical}, and representative Single-Source DG Approaches \textbf{RSC}~\cite{huang2020self} and \textbf{SD}~\cite{pezeshki2021gradient} to finetune models and compute the estimation of CI Index.

\subsection*{Appendix B.3: Creation of pairs}
In the definition of CI Index \ref{eq.dg-index}, it compute the JSD of sample pairs with identical $x^{c}$ and varied $x^{n}$. In this experiment, we pick 5\% of samples in domain \emph{only-fg} as pure set, which only contain objects mapping to the ground-truth label. We can regard them as approximation of $x^{c}$. For $x^{n}$, we can put samples from \emph{only-fg} onto randomly selected samples in domain \emph{original}. When estimating the CI Index, we iterate the pure set 20 times to compute the average of JSD.

\subsection*{Appendix B.4: Experimental setting}
We conduct the experiments on ViT-B/16~\cite{dosovitskiy2020image}, with CLIP pre-training weights \cite{radford2021learning}.
We adopt the training strategy (LP-FT) proposed in \cite{kumar2022fine} to more effectively reuse the learned representations of the pre-trained weights.
In short, LP-FT consists of linear probing (LP) on the linear classifier while freezing the feature extractor, followed by fine-tuning (FT) the entire model. 

For LP, we trained 10 epochs. Learning rate is 0.003. For FT, we trained 20 epochs Learning rate is 3e-5. 
After training, we estimate the CI Index for these models on pure set. The results are shown in
Section \ref{sec.pcip}.

\end{document}